%% file: example.tex
\documentclass{article}
\usepackage{subcaption}
\usepackage[ruled, vlined, linesnumbered]{algorithm2e}
\SetKwFunction{Fn}{Fn}
\usepackage{amssymb}
\usepackage{amsmath}
\usepackage{graphicx}
\usepackage{verbatim}
\usepackage{multirow}
\usepackage{placeins}
\usepackage{multirow}
\usepackage{setspace}
\usepackage[utf8]{inputenc} 
\usepackage[T1]{fontenc}    
\usepackage{hyperref}       
\usepackage{url}            
\usepackage{booktabs}       
\usepackage{amsfonts}       
\usepackage{nicefrac}       
\usepackage{microtype}      
\usepackage{lipsum}
\usepackage{graphicx}
\graphicspath{{media/}}     
\usepackage{url}
\usepackage{hyperref}
\usepackage{algorithm2e}
\usepackage{wrapfig}
\usepackage{caption}
\usepackage{float}
\usepackage{tcolorbox}
\usepackage{ragged2e}
\SetAlFnt{\small}
\SetAlCapNameFnt{\small}
\SetAlCapFnt{\small}
\SetFuncSty{textsc}
\AtBeginEnvironment{algorithm}{
	\DontPrintSemicolon
}
\usepackage{booktabs}

\usepackage{PRIMEarxiv}

\usepackage[utf8]{inputenc} 
\usepackage[T1]{fontenc}    
\usepackage{hyperref}       
\usepackage{url}            
\usepackage{booktabs}       
\usepackage{amsfonts}       
\usepackage{nicefrac}       
\usepackage{microtype}      
\usepackage{lipsum}
\usepackage{fancyhdr}       
\usepackage{graphicx}       
\graphicspath{{media/}}     

\usepackage{PRIMEarxiv}
\newcommand\pimodel{$\pi_0$} 
\title{FATE-VLA: Failure-Aware Test Generation for Vision-Language-Action Models}

\author{
   Arusa Kanwal\\
   Mondragon University \\
   Mondragon, Spain\\
   \texttt{akanwal@mondragon.edu} 
   \And
   Pablo Valle\\
   Mondragon University \\
   Mondragon, Spain \\
   \texttt{pvalle@mondragon.edu} \\
  \And
   Shaukat Ali\\
   Simula Research Laboratory \\
   Oslo, Norway \\
   \texttt{shaukat@simula.no} \\
\And
   Aitor Arrieta\\
   Mondragon University \\
   Mondragon, Spain \\
   \texttt{aarrieta@mondragon.edu}
}

\begin{document}
\maketitle
\begin{abstract}
Vision–Language–Action (VLA) models are increasingly used as generalist robot policies, yet their evaluation still relies largely on static benchmarks that randomly sample task scenes. In high-dimensional embodied spaces, failures are sparse and clustered, so static benchmarking can underestimate robustness risks. We reframe VLA evaluation as an active failure-discovery problem and propose a failure-aware test-generation approach that combines diversity-driven exploration with surrogate models learned from observed executions. The method steers testing toward high-risk yet diverse scene regions. Across four state-of-the-art VLA models, it uncovers substantially more failures (up to +29.7 \% over selected baselines) while revealing more diverse failure modes. This means that, for instance, in the case of GR00T-N1.6, success rate dropped from 64.4\% to 34.7\%. More broadly, our findings call for a shift in VLA evaluation: from passive measurement on fixed task suites to adaptive, failure-seeking test generation that exposes the structure of model weaknesses before deployment.

\end{abstract}


\vspace{-2mm}
\section{Introduction}
\vspace{-3mm}
Vision-Language-Action (VLA) models are emerging as a promising paradigm for generalist robot policies, enabling robots to interpret natural-language instructions and execute manipulation or navigation behaviors. Initiatives such as Open X-Embodiment \cite{openx2023} and open foundation models such as OpenVLA \cite{kim2024openvla} have accelerated this progress, while benchmarks including CALVIN~\cite{calvin2022}, LIBERO~\cite{libero2023}, and ManiSkill2~\cite{maniskill22023} provide standardized environments for evaluating VLA models. Despite these advances, VLA evaluation remains largely based on success rates over predefined or randomly sampled task scenes. Such static benchmarking is limited in high-dimensional embodied interaction spaces, where scene configurations vary across object identities, spatial layouts, orientations, and instruction semantics. Failures in such spaces are often sparse, non-uniform, and clustered in specific regions. Consequently, naive sampling may underestimate risk and overlook behavioral fragilities.

Moreover, effective evaluation should consider not only how many failures are found, but also how diverse those failures are. Repeatedly exposing the same weakness under near-identical scene configurations provides limited diagnostic value, whereas failures spanning different objects, spatial arrangements, and behavioral trajectories reveal a broader picture of model vulnerabilities. Recent work, such as VLATest~\cite{vlatest2024}, has highlighted the importance of robustness and behavioral diversity in VLA evaluation. However, existing approaches typically rely on random sampling or predefined perturbations, without explicitly learning where failure-prone regions lie or adaptively steering test generation toward them.

We therefore argue that VLA evaluation should be reframed as an \emph{active failure discovery problem}: rather than passively measuring performance on static benchmarks, testing should progressively learn the structure of model weaknesses and generate scenarios that are both failure-prone and diverse. To this end, we propose FATE-VLA, a Failure-Aware Test Generation approach that combines diversity-driven exploration inspired by Adaptive Random Testing (ART)~\cite{chen2004adaptive} with surrogate models that estimate the likelihood of failure. By iteratively learning from previously executed scenes, our approach guides test generation toward high-risk yet diverse regions of the scene space.

Beyond failure rate, we introduce trajectory-level and object-level failure diversity metrics to characterize the heterogeneity of discovered weaknesses. We evaluate our approach on four state-of-the-art VLA models and show that failure-aware generation uncovers substantially more failures, as well as more diverse failure modes, than random and diversity-only baselines. These findings support a shift from static benchmark evaluation toward adaptive, diagnostic testing pipelines that can expose systematic robustness gaps before VLA models are deployed on physical robots.

This paper makes the following contributions: (1) We formulate VLA testing as an \emph{active failure discovery problem}, emphasizing the joint importance of failure detection and failure diversity; (2) We propose FATE-VLA, a failure-aware test generation approach that combines diversity-driven exploration with surrogate-assisted failure prediction; (3) We introduce trajectory-level and object-level failure diversity metrics to characterize discovered weaknesses beyond conventional success rate; (4) We empirically evaluate the approach on four state-of-the-art VLA models, showing that it uncovers more failures and more diverse failure modes than random and diversity-driven baselines. In addition, we provide a complete replication package along with all the results from our experiments \cite{FATE-VLA_code}.\vspace{-2mm}

\section{Related Work}
\vspace{-3mm}
Recent VLA models, from RT-1~\cite{brohan2022rt} to OpenVLA~\cite{kim2024openvla}, SpatialVLA~\cite{qu2025spatialvla}, GR00T~\cite{nvidia2025gr00tn1openfoundation}, $\pi_0$~\cite{black2024pi0visionlanguageactionflowmodel}, and EO-1~\cite{qu2025eo1}, have advanced robotic manipulation by mapping language and visual observations to low-level actions. However, their robustness remains insufficiently understood. Existing studies evaluate VLA robustness through perturbation- or adversarial-based methods: VLATest~\cite{wang2025vlatest} studies variations in lighting, viewpoints, object configurations, and instructions; Eva-VLA~\cite{liu2025eva}, Nebula~\cite{peng2025nebula}, and DAERT~\cite{tong2026uncovering} seek worst-case or adversarial failures; and benchmarks such as VLABench~\cite{zhang2025vlabench}, LIBERO-Plus~\cite{fei2025libero}, and LIBERO-X~\cite{wang2026libero} reveal brittleness under distribution shifts, sensor noise, and language variation. Yet these approaches largely rely on predefined perturbations or adversarial objectives rather than adaptively discovering diverse failure cases.

Related work has also explored failure prediction and failure discovery in robotic and safety-critical systems. Recent studies~\cite{gu2026safe,farid2022failure} focus on predicting failures, while others~\cite{lee2020adaptive,haq2022efficient} actively search for failure-inducing scenarios. However, these approaches are not designed for multimodal VLA evaluation. Similarly, Valle et al.~\cite{valle2026metamorphic} use metamorphic testing to detect inconsistent VLA behaviors, THE COLOSSEUM~\cite{pumacay2024colosseum} provides a robustness benchmark, CPS-Fuzz~\cite{sheikhi2022coverage} applies coverage-guided fuzzing to cyber-physical systems, and RoboArena~\cite{atreya2025roboarena} supports large-scale policy ranking. These works emphasize evaluation, coverage, or prediction, but not targeted discovery of diverse failure regions in VLA systems through methods that generate test scenes in an automated manner.

In contrast, we formulate VLA evaluation as an active failure discovery problem. Our approach explicitly learns failure-prone regions and adaptively generates test scenarios that expose diverse weaknesses across object configurations, spatial layouts, and behavioral trajectories.
\vspace{-2mm}

\section{Test Generation Algorithm}
\vspace{-3mm}
We propose two hybrid test-generation algorithms that combine Adaptive Random Testing (ART) ~\cite{chen2004adaptive} with machine-learning guidance to discover diverse, failure-inducing VLA scenes. The first augments FSCS-ART with a surrogate model that prioritizes candidates predicted to fail in subsection ~\ref{sec:algo1}, while the second uses Random Forest failure probabilities within an adaptive exploration–exploitation scoring function in subsection ~\ref{sec:algo2}. Together, they aim to improve failure discovery without sacrificing test diversity.
\vspace{-2mm}
\subsection{Algorithm 1 - Hybrid Sorted Surrogate Assisted FSCS-ART Testing Algorithm}
\label{sec:algo1}
The first algorithm augments FSCS-ART~\cite{chen2004adaptive} with surrogate-guided failure prediction. While FSCS-ART promotes diversity through distance-based candidate selection, it does not explicitly target failure-prone regions. We therefore iteratively train a surrogate model on scenes that are being executed through the algorithm to prioritize candidates likely to expose failures while preserving input diversity. This is well-suited to VLA testing, where scene spaces defined by object type, position, and orientation are too large for exhaustive exploration.
Algorithm~\ref{alg:hybrid} explains our procedure. The algorithm takes as inputs the VLA agent under test $VLA_{UT}$, an object pool $O$, spatial constraints defined by $(X_{range}, Y_{range})$, orientation parameters $\Omega$, the candidate set size $k$, and the total number of test scenes $N$ to be generated. During execution, two archives are maintained. The first is the tested scene archive $S$, which stores all executed scenes. The second is the labeled archive $D$, which records the outcome of each executed scene with scene details, i.e., $D = \{(s_i, f_i)\}$, where $s_i$ represents a scene (object name, location and orientation) and $f_i \in \{0,1\}$ denotes the observed outcome ($1$ indicates failure and $0$ indicates success). Initially, both $S$ and $D$ are empty, and no machine learning model (lines~\ref{algohybrid_2} - \ref{algohybrid_4}) is available due to the lack of labeled data. The testing process continues iteratively until the $N$ scenes have been executed. In each iteration, a candidate set $C$ of size $k$ (line~\ref{algohybrid_8}), i.e., $C = \{c_1, c_2, ..., c_k\}$ is generated using random scene generation subject to the defined spatial and object constraints.
\begin{wraptable}{r}{0.40\textwidth}
\vspace{-1mm}
\begin{minipage}{0.40\textwidth}
\begin{tcolorbox}[
colback=white,
colframe=white,
boxrule=0.35pt,
sharp corners,
left=0.5mm,
right=0.5mm,
top=0.3mm,
bottom=0.3mm
]
\vspace{-2mm}
\begin{algorithm}[H]
\tiny
\setstretch{0.72}
\DontPrintSemicolon
\SetAlgoSkip{0pt}
\SetInd{0.15em}{0.3em}
\SetKwInOut{Input}{input}
\SetKwInOut{Output}{output}
\Input{
$VLA_{UT}$: Vision-Language Agent Under Test,
$O$: Object pool,
$(X_{range}, Y_{range})$: Position bounds,
$\Omega$: Orientation set,
$k$: Candidate set size,
$N$: Number of test scenes
}
\Output{
$\mathcal{S}$: Executed scene archive,
$\mathcal{D}$: Dataset,
$DT$: Trained ML Models
}
{
$\mathcal{D} \leftarrow \varnothing$\tcp*{Archive} \label{algohybrid_2}
$\mathcal{S} \leftarrow \varnothing$\tcp*{Tested scene archive} \label{algohybrid_3}
$DT \leftarrow \text{None}$\; \label{algohybrid_4}
$i \leftarrow 0$\; \label{algohybrid_5}
\While{$i < N$}{ \label{algohybrid_5}
    $i \leftarrow i + 1$\; \label{algohybrid_7}
    $C \leftarrow$ \textsc{GenerateRandomScenes}$(k, O, X_{range}, Y_{range}, \Omega)$\; \label{algohybrid_8}
    \eIf{$DT = \text{None}$}{ \label{algohybrid_9}
        $s \leftarrow$ \textsc{FSCS\_ART\_Select}$(C, \mathcal{S})$ \label{algohybrid_10} \tcp*{Exploration} 
    }{
        $C_{sorted} \leftarrow$ \textsc{SortByMaxMinDistance}$(C, \mathcal{S})$\; \label{algohybrid_12}
        $s \leftarrow \text{None}$\; \label{algohybrid_13}
        \ForEach{$c \in C_{sorted}$}{ \label{algohybrid_14}
            \If{\textsc{DT\_Predict}$(c) = 1$}{ \label{algohybrid_15}
                $s \leftarrow c$\tcp*{Predicted failure} \label{algohybrid_16}
                \textbf{break}\;
            }
        }
        \If{$s = \text{None}$}{ \label{algohybrid_18}
            $s \leftarrow$ \textsc{First}$(C_{sorted})$ \label{algohybrid_19}
        }
    }
    $f \leftarrow$ \textsc{ExecuteAndObserve}$(s, VLA_{UT})$\tcp*{$1$=failure, $0$=success} \label{algohybrid_20}
    $\mathcal{D} \leftarrow \mathcal{D} \cup \{(s,f)\}$\; \label{algohybrid_21}
    $\mathcal{S} \leftarrow \mathcal{S} \cup \{s\}$\; \label{algohybrid_22}
    \If{$\mathcal{D}$ contains at least two classes}{ \label{algohybrid_23}
        $DT \leftarrow$ \textsc{TrainML}$(\mathcal{D})$ \label{algohybrid_24}
    }
}
\Return{$\mathcal{S}, \mathcal{D}$}\; \label{algohybrid_25}
}
\caption{\tiny Hybrid Sorted Surrogate Assisted FSCS-ART}
\label{alg:hybrid}
\end{algorithm}
\end{tcolorbox}
\end{minipage}
\end{wraptable}
\justifying
During the initial exploration phase (lines~\ref{algohybrid_9}-~\ref{algohybrid_10}), when no trained surrogate model is available, the algorithm relies on diversity-based sampling rather than learned guidance. For each candidate $c \in C$, the minimum distance to previously executed scenes in $S$ is computed using the Euclidean distance metric, $dist(c,s) = \sqrt{\sum_{j=1}^{m}(c_j - s_j)^2}$, where $m$ denotes the number of scene features. Euclidean distance is used because scene parameters are represented as continuous numerical feature vectors, making it a simple and computationally efficient measure of dissimilarity. The candidate selected for execution is the one that maximizes the minimum distance from previously executed scenes, i.e., $s^* = \arg\max_{c \in C} \left( \min_{s \in S} dist(c,s) \right)$. This selection mechanism promotes diversity among test inputs, reduces redundancy, and encourages broad exploration of the input space.

After executing the selected scene on the VLA model, the observed outcome is recorded and the archives are updated such that $S \leftarrow S \cup \{s^*\}$ and $D \leftarrow D \cup \{(s^*, f)\}$ (Lines~\ref{algohybrid_20}-~\ref{algohybrid_22}). As execution proceeds, the archive gradually accumulates labeled examples describing both successful and failure-inducing scenes. Once the archive contains examples from both outcome classes (success and failure), a surrogate learning model $M$ is trained using the archive $D$ (line~\ref{algohybrid_24}). The model learns a mapping between scene features and predicted outcomes, i.e., $\hat{f} = M(s)$, where $\hat{f}$ denotes the predicted failure label for scene $s$. In subsequent iterations (Lines~\ref{algohybrid_12}-~\ref{algohybrid_19}), candidate scenes are first sorted according to their diversity with respect to $S$, preserving the exploration capability of FSCS-ART. The trained model then evaluates each candidate and predicts whether it is likely to cause a failure. If any candidate is predicted as a failure, the algorithm prioritizes the first such candidate for execution. Otherwise, the candidate with the highest diversity score is selected.
After each execution, the archive is updated and the model retrained to incorporate newly observed data. This iterative learning process progressively improves the model’s ability to identify failure-prone regions of the input space.

The hybrid strategy balances \emph{exploration} and \emph{exploitation}. FSCS-ART ensures exploration by maximizing diversity among test scenes, which increases coverage of the input domain. The surrogate model provides exploitation by predicting regions with a higher probability of failure based on previously observed outcomes. By combining these two mechanisms, the algorithm increases the likelihood of discovering failures while aiming to maximize the input space to discover a broad and diverse set of failures.
\subsection{Algorithm 2 - Hybrid Adaptive RF-Guided FSCS-ART Testing Algorithm}
\label{sec:algo2}
The Hybrid Adaptive RF–Guided FSCS-ART Testing Algorithm (\ref{alg:hybrid_rf}) is designed to automatically generate and execute test scenes for evaluating a VLA model. Similarly to Algorithm~\ref{alg:hybrid}, the algorithm takes as input the VLA under test $VLA_{UT}$, an object pool $O$, spatial limits $(X_{range}, Y_{range})$ defining possible object positions, an orientation set $\Omega$, the size of the candidate set $k$, and the total number of test executions $N$. During execution, the algorithm maintains three key structures (Lines~\ref{algohybrid_rf_2}-~\ref{algohybrid_rf_4}): an archive $D$ containing executed test scenes and their outcomes, an executed scene archive $S$ storing all previously executed tests, and a failure region archive $F$ storing scenes that resulted in failures. Initially, the RF model $RF$ is untrained (Line~\ref{algohybrid_rf_5}); therefore, the algorithm begins with an exploration-based strategy to collect initial execution data before surrogate-guided test generation is applied.

In the initial exploration phase (Lines~\ref{algohybrid_rf_9} -~\ref{algohybrid_rf_11}), the algorithm generates a set of candidate test scenes $C$ from the object pool and the defined spatial parameter ranges, including possible object positions within $(X_{range}, Y_{range})$ and orientations from $\Omega$. The candidate set can be represented as $C = {c_1, c_2, \dots , c_k}$, where each candidate $c_i$ represents a test configuration consisting of object selection, spatial positions, and orientation parameters. The FSCS-ART strategy selects the candidate that is maximally distant from previously executed scenes in $S$. The distance between two scenes $c$ and $s$ is computed using a distance function $Dist(c,s)$. The selected scene is determined as $s = \arg\max_{c \in C} \left( \min_{s' \in S} Dist(c, s) \right)$.
This selection ensures that the next test case is as different as possible from previously executed scenes, encouraging diversity and broader exploration of the input space. The selected scene $s$ is then executed on the VLA model, producing a FAIL/SUCCESS outcome $f$. The executed scene and its result are stored in the archive $D$.
When the archive $D$ contains both successful and failed cases, the algorithm enters the learning-guided phase (Lines~\ref{algohybrid_rf_13}-~\ref{algohybrid_rf_24}). 
\begin{wrapfigure}{r}{0.40\textwidth}
\begin{minipage}{0.40\textwidth}
\begin{tcolorbox}[
colback=white,
colframe=white,
boxrule=0.35pt,
sharp corners,
left=0.5mm,
right=0.5mm,
top=0.3mm,
bottom=0.3mm
]
\vspace{-2mm}
\begin{algorithm}[H]
\tiny
\setstretch{0.72}
\DontPrintSemicolon
\SetAlgoSkip{0pt}
\SetInd{0.15em}{0.3em}
\SetKwInOut{Input}{input}
\SetKwInOut{Output}{output}
\Input{
$VLA_{UT}$: Vision-Language Agent Under Test,
$O$: Object pool,
$(X_{range}, Y_{range})$: Position bounds,
$\Omega$: Orientation set,
$k$: Candidate set size,
$N$: Number of test scenes
}
\Output{
$\mathcal{S}$: Executed scene archive,
$\mathcal{D}$: Dataset,
$RF$: Trained Random Forest model
}
\vspace{2mm}
{
$\mathcal{D} \leftarrow \varnothing$\tcp*{Archive} \label{algohybrid_rf_2}
$\mathcal{S} \leftarrow \varnothing$\tcp*{Executed scene archive} \label{algohybrid_rf_3}
$\mathcal{F} \leftarrow \varnothing$\tcp*{Failure region  archive} \label{algohybrid_rf_4}
$RF \leftarrow \text{None}$\; \label{algohybrid_rf_5}
$i \leftarrow 0$\; \label{algohybrid_rf_6}
\While{$i < N$}{ \label{algohybrid_rf_7}
    $i \leftarrow i + 1$\; \label{algohybrid_rf_8}
    \eIf{$RF = \text{None}$}{ \label{algohybrid_rf_9}
        $C \leftarrow$ \textsc{GenerateRandomScenes}$(k, O, X_{range}, Y_{range}, \Omega)$\; \label{algohybrid_rf_10}
        $s \leftarrow$ \textsc{FSCS\_ART\_Select}$(C, \mathcal{S})$\tcp*{Exploration phase} \label{algohybrid_rf_11}
    }{
        $C \leftarrow$ \textsc{GenerateCandidates}$(\mathcal{F}, k)$\tcp*{Local + global mix} \label{algohybrid_rf_13}
        $\alpha \leftarrow \min(0.85, \frac{i}{N})$\tcp*{Adaptive weight} \label{algohybrid_rf_14}
        $best\_score \leftarrow -\infty$\; \label{algohybrid_rf_15}
        $s \leftarrow \text{None}$\; \label{algohybrid_rf_16}
        \ForEach{$c \in C$}{ \label{algohybrid_rf_17}
            $d \leftarrow \min_{s' \in \mathcal{S}} \textsc{Dist}(c, s')$\; \label{algohybrid_rf_18}
            $norm\_d \leftarrow \textsc{Normalize}(d)$\; \label{algohybrid_rf_19}
            $p_{fail} \leftarrow RF.\textsc{PredictProb}(c)$\; \label{algohybrid_rf_20}
            $score \leftarrow \alpha \cdot p_{fail} 
            + (1-\alpha) \cdot norm\_d$\; \label{algohybrid_rf_21}
            \If{$score > best\_score$}{ \label{algohybrid_rf_22}
                $best\_score \leftarrow score$\; \label{algohybrid_rf_23}
                $s \leftarrow c$\; \label{algohybrid_rf_24}
            }
        }
    }
    $f \leftarrow$ \textsc{ExecuteAndObserve}$(s, VLA_{UT})$\tcp*{$1$=failure, $0$=success} \label{algohybrid_rf_25}
    $\mathcal{D} \leftarrow \mathcal{D} \cup \{(s,f)\}$\; \label{algohybrid_rf_26}
    $\mathcal{S} \leftarrow \mathcal{S} \cup \{s\}$\; \label{algohybrid_rf_27}
    \If{$f = 1$}{ \label{algohybrid_rf_28}
        $\mathcal{F} \leftarrow \mathcal{F} \cup \{s\}$\tcp*{Store failure region} \label{algohybrid_rf_29}
    }
    \If{$\mathcal{D}$ contains at least two classes}{ \label{algohybrid_rf_30}
        $RF \leftarrow$ \textsc{TrainRandomForest}$(\mathcal{D})$\; \label{algohybrid_rf_31}
    }
}
\Return{$\mathcal{S}, \mathcal{D}, RF$}\; \label{algohybrid_rf_32}
}
\caption{\tiny Hybrid Adaptive RF-Guided FSCS-ART Testing Framework}
\label{alg:hybrid_rf}
\end{algorithm}
\end{tcolorbox}
\end{minipage}
\end{wrapfigure}
\justifying
In this phase, candidate scenes are generated using the failure archive $F$, enabling the algorithm to explore regions near previously discovered failures while still maintaining global diversity. For each candidate scene $c$, the minimum distance to previously executed tests is computed as $d = \min_{s \in S} Dist(c, s)$ and normalized to the range $[0,1]$ using $norm_d = \frac{d - d_{min}}{d_{max} - d_{min}}$ to ensure comparability across candidates. This normalized distance represents the diversity contribution of the candidate and encourages exploration of new regions in the test space. Simultaneously, the trained surrogate model, implemented using a RF model, estimates the failure probability of each candidate as $p_{fail} = RF.predictProb(c)$ (Line~\ref{algohybrid_rf_20}). This probability reflects the surrogate model’s learned knowledge of configurations that are more likely to produce failures based on the previously observed execution data stored in $D$. To balance failure prediction and exploration diversity, a combined score (Lines ~\ref{algohybrid_rf_21}-~\ref{algohybrid_rf_24}) is calculated for each candidate by using $score(c) = \alpha \cdot p_{fail} + (1-\alpha) \cdot norm_d$, where $\alpha$ is an adaptive weight controlling the relative importance of failure prediction versus diversity. The candidate with the highest score is selected for execution, i.e., $s = \arg\max_{c \in C} score(c)$.
This mechanism prioritizes scenes that are both likely to reveal failures and sufficiently distinct from previously executed tests, ensuring both, diversity of the generated tests and effectiveness.

The selected scene is executed on the VLA model (Line~\ref{algohybrid_rf_25}). If the scene results in failure ($f = 1$), it is added to the failure archive $F$ (Lines~\ref{algohybrid_rf_28}, \ref{algohybrid_rf_29}). All executed scenes and their outcomes are continuously stored in the archive $D$ (Line~\ref{algohybrid_rf_26}). The surrogate model is retrained at each iteration with $D$ (Line~\ref{algohybrid_rf_31}), incrementally improving its ability to predict scenes that are likely to lead to failures. This iterative process continues until $N$ test scenes have been executed. By combining adaptive random testing with machine learning–based guidance used as a surrogate model, the algorithm effectively balances exploration and exploitation, improving the likelihood of discovering failure-inducing scenes in the VLA model.
\vspace{-2 em}
\section{Experiment Design}
The goal of our experimental evaluation is to answer the following Research Questions (RQ): 

\textbf{RQ1 -- \textit{How does our approach perform compared to the selected baselines?}} This RQ evaluates the effectiveness of the proposed approach by comparing it against representative baselines, including random and diversity-driven strategies. It also verifies that the problem is sufficiently challenging to justify the use of a more sophisticated, learning-guided technique. 

\textbf{RQ2 -- \textit{How do different surrogate models and the two algorithms help in the test generation process?}} We aim to analyze how different surrogate models and test generation algorithms influence the effectiveness of the proposed approach. It investigates how each component contributes to guiding the search process, shaping the exploration–exploitation balance, and ultimately improving failure detection and diversity.
\paragraph{Subject VLAs and Simulation Benchmark:} 
Our evaluation was conducted with four state-of-the-art VLA models: OpenVLA~\cite{kim2024openvla}, \pimodel~\cite{black2024pi0visionlanguageactionflowmodel}, GR00T-N1.6~\cite{nvidia2025gr00tn1openfoundation}, and EO-1~\cite{qu2025eo1}. We employed the widely used SimplerEnv~\cite{li24simpler} simulation environment, following recent testing studies~\cite{valle2026metamorphic, valle2025evaluating, wang2025vlatest}. We only used the \textit{``Pick up''} task because it is both (1) simple and (2) the one on which the models showed the highest performance in related studies~\cite{valle2026metamorphic, valle2025evaluating, wang2025vlatest}. Subsequently, identifying failures in such a task is more challenging. Following related studies~\cite{valle2026metamorphic, valle2025evaluating, wang2025vlatest}, the task is considered successful if the target object has been successfully grasped by the robot and lifted.
\vspace{-1mm}

\paragraph{Number and type of objects:} The failure-aware test generation approach uses seven different manipulation objects to create diverse robotic testing scenarios, including a RedBull can, blue plastic bottle, eggplant, carrot, apple, spoon, and sponge. For each generated scene, one object is randomly selected and positioned within the predefined workspace boundaries of SimplerEnv, where the $x$-coordinate ranges from $[-0.5, -0.05]$ and the $y$-coordinate ranges from $[0.0, 0.4]$.
\vspace{-1mm}

\paragraph{Evaluation Metrics:} To evaluate the different approaches, we consider four evaluation metrics: 
\textbf{(i) Failure Rate (FR):} Measures the proportion of generated test scenes that lead to unsuccessful task execution by the VLA under test. Formally, given a set of executed scenes $S$, the FR is defined as: $FR = \frac{|S_{fail}|}{|S|}$, where $S_{fail} \subseteq S$ denotes the subset of scenes in which the VLA fails to successfully complete the instructed task. 
\textbf{(ii) Trajectory Coverage (TC):} Following the trajectory-based adequacy notion introduced in VLATest~\cite{vlatest2024}, this metric quantifies the extent to which the generated test scenes explore novel target object positions, i.e., the object positions covered with regards to the size of the manipulation platform (i.e., the desk).
Trajectory coverage is defined as the ratio of visited positions with respect to the total number of reachable positions: $TC = \frac{|P_{visited}|}{|P|}$, where $P_{visited}$ denotes the set of discretized position regions exercised by the test suite and $P$ represents the full reachable position space. 
\textbf{(iii) Trajectory Coverage under Failure (TCF):} This metric measures trajectory coverage considering only those executions that resulted in task failure, providing a quantitative estimate of the behavioral diversity of the discovered failures: $TCF = \frac{|B_{fail}|}{|B|}$, where $B_{fail}$ denotes the set of trajectory bins exercised by failing executions. Higher values indicate that failures occur across heterogeneous behavioral regions rather than being concentrated around a single execution pattern. 
\textbf{(iv) Failed Object Coverage (FOC):} It measures the proportion of distinct manipulated objects for which at least one failure was observed. Let $O$ denote the set of objects involved in the evaluation and $O_{fail} \subseteq O$ the subset for which at least one test scene resulted in failure, i.e., FOC = $\frac{|O_{fail}|}{|O|}$. This metric captures the extent to which the testing technique uncovers object-specific failure modes, which are common in embodied manipulation tasks.
\vspace{-1mm}
\paragraph{Employed ML models as surrogate models:} We selected Decision Tree (DT) and Random Forest (RF) as classifiers since they are lightweight, computationally efficient, and suitable for iterative retraining during adaptive testing. DT provides simple and interpretable decision boundaries for identifying failure-prone regions, while RF improves prediction robustness and generalization through ensemble learning. 
\vspace{-1mm}
\paragraph{Hyperparameters:}
An RF classifier was used to predict failure-prone scenes from object type, position, and orientation features, using 150 trees, a maximum depth of 12, and a minimum split size of 3. A DT classifier was also evaluated with a maximum tree depth of 8 to control model complexity and reduce overfitting. Candidate scenes were ranked using predicted failure probability and diversity distance through an adaptive weighting parameter, $\alpha = \min(0.85, \frac{i}{N})$, which gradually shifted the search from exploration to exploitation while preserving some diversity.
\vspace{-1 em}
\paragraph{Baseline Algorithms:}We considered two baseline algorithms: (1) a purely random algorithm and (2) an Adaptive Random Testing (ART) algorithm without the use of a DT. The random algorithm enables comparison against a standard baseline, particularly since we argue that the aspects addressed by our approach are not explicitly considered in state-of-the-art VLA testing approaches. The ART variant allows us to conduct an ablation study to assess the impact of incorporating the DT component on the overall performance of our approach.
\vspace{-1mm}
\paragraph{Execution Runs:} Each experimental configuration was executed 10 times to account for randomness and ensure statistically meaningful comparisons. 
\vspace{-2mm}
\section{Experimental Results}
\vspace{-3mm}
\label{sec:result}
Table~\ref{tab:summary} summarizes the best proposed result per VLA and metric, alongside both baseline values for direct comparison, while detailed results can be found in Appendix~\ref{sec:appendix_tables}. Across all non-ceiling VLAs, the proposed methods improve FR by 14--30~\% over B1 and 13--29~\% over B2, while maintaining or improving TC, TCF, and FOC. \texttt{Sorting\_RF} is the strongest performer on FR and TCF for EO1 and GR00T-N1.6; \texttt{Weighted\_RF} leads on FR for \pimodel \space and on FOC for EO1 and GR00T-N1.6. RQ2 analysis shows that the ML model is the primary driver of improvement, the near-zero gap between B1 and B2 confirms that ART diversity alone adds nothing to failure detection, while adding the ML model (DT or RF) creates the full observed advantage.\vspace{-6mm}
\begin{table}[!ht]
\centering
\caption{Summary of best proposed algorithm per metric and VLA ($\uparrow$ = higher is better), with gains over both baselines.}
\label{tab:summary}
\resizebox{\linewidth}{!}{%
\begin{tabular}{llccccc}
\toprule
\textbf{VLA} & \textbf{Metric} & \textbf{B1: Random} & \textbf{B2: FSCSART} & \textbf{Best Proposed} & \textbf{Algorithm} & \textbf{$\Delta$B1 / $\Delta$B2} \\
\cmidrule{1-7}
\multirow{4}{*}{\textbf{EO1}}
  & FR  (\% )$\uparrow$ & 36.7  & 38.0  & 60.0  & \texttt{Sorting\_RF}  & +23.3 / +22.0~\% \\
  & TC  (\% )$\uparrow$ & 82.8  & 81.5  & 84.0  & \texttt{Sorting\_DT}  & +1.2 / +2.5~\%   \\
  & TCF (\% )$\uparrow$ & 70.8  & 70.3  & 72.2  & \texttt{Sorting\_DT}  & +1.4 / +1.9~\%   \\
  & FOC (\% )$\uparrow$ & 96.1  & 98.6  & 98.6  & \texttt{Weighted\_RF} & +2.5 / 0.0~\%    \\
\cmidrule{1-7}
\multirow{4}{*}{\textbf{GR00T-N1.6}}
  & FR  (\% )$\uparrow$ & 35.6  & 36.5  & 65.3  & \texttt{Sorting\_RF}  & +29.7 / +28.8~\% \\
  & TC  (\% )$\uparrow$ & $\sim$86 & $\sim$86 & $\sim$86 & ---          & $\approx$0~\%    \\
  & TCF (\% )$\uparrow$ & 78.2  & 79.9  & 83.0  & \texttt{Sorting\_RF}  & +4.8 / +3.1~\%   \\
  & FOC (\% )$\uparrow$ & 95.7  & 95.8  & 97.1  & \texttt{Weighted\_RF} & +1.4 / +1.3~\%   \\
\cmidrule{1-7}
\multirow{4}{*}{\textbf{OpenVLA-7b}}
  & FR  (\% )$\uparrow$ & 95.3  & 95.7  & 97.5  & \texttt{Weighted\_RF} & +2.2 / +1.8~\%   \\
  & TC  (\% )$\uparrow$ & 66.8  & 67.8  & 70.5  & \texttt{Sorting\_RF}  & +3.7 / +2.7~\%   \\
  & TCF (\% )$\uparrow$ & 63.2  & 65.4  & 70.1  & \texttt{Sorting\_RF} / \texttt{Weighted\_RF} & +6.9 / +4.7~\% \\
  & FOC (\% )$\uparrow$ & 100.0 & 100.0 & 100.0 & ---                   & 0.0 / 0.0~\%     \\
\cmidrule{1-7}
\multirow{4}{*}{$\boldsymbol{\pi_0}$}
  & FR  (\% )$\uparrow$ & 54.0  & 55.5  & 73.7  & \texttt{Weighted\_RF} & +19.7 / +18.2~\% \\
  & TC  (\% )$\uparrow$ & 82.8  & 83.2  & 84.2  & \texttt{Sorting\_DT}  & +1.4 / +1.0~\%   \\
  & TCF (\% )$\uparrow$ & 72.1  & 72.6  & 77.4  & \texttt{Sorting\_RF}  & +5.3 / +4.8~\%   \\
  & FOC (\% )$\uparrow$ & 95.7  & 100.0 & 100.0 & \texttt{Sorting\_RF} / \texttt{Weighted\_RF} & +4.3 / 0.0~\% \\
\bottomrule
\end{tabular}}
\end{table}
\vspace{4mm}
\subsection{RQ1: Performance Against Both Baselines}
\label{sec:rq1}
\begin{figure}[htbp]
    \centering
    \begin{subfigure}[b]{0.48\linewidth}
        \centering
        \includegraphics[width=\linewidth]{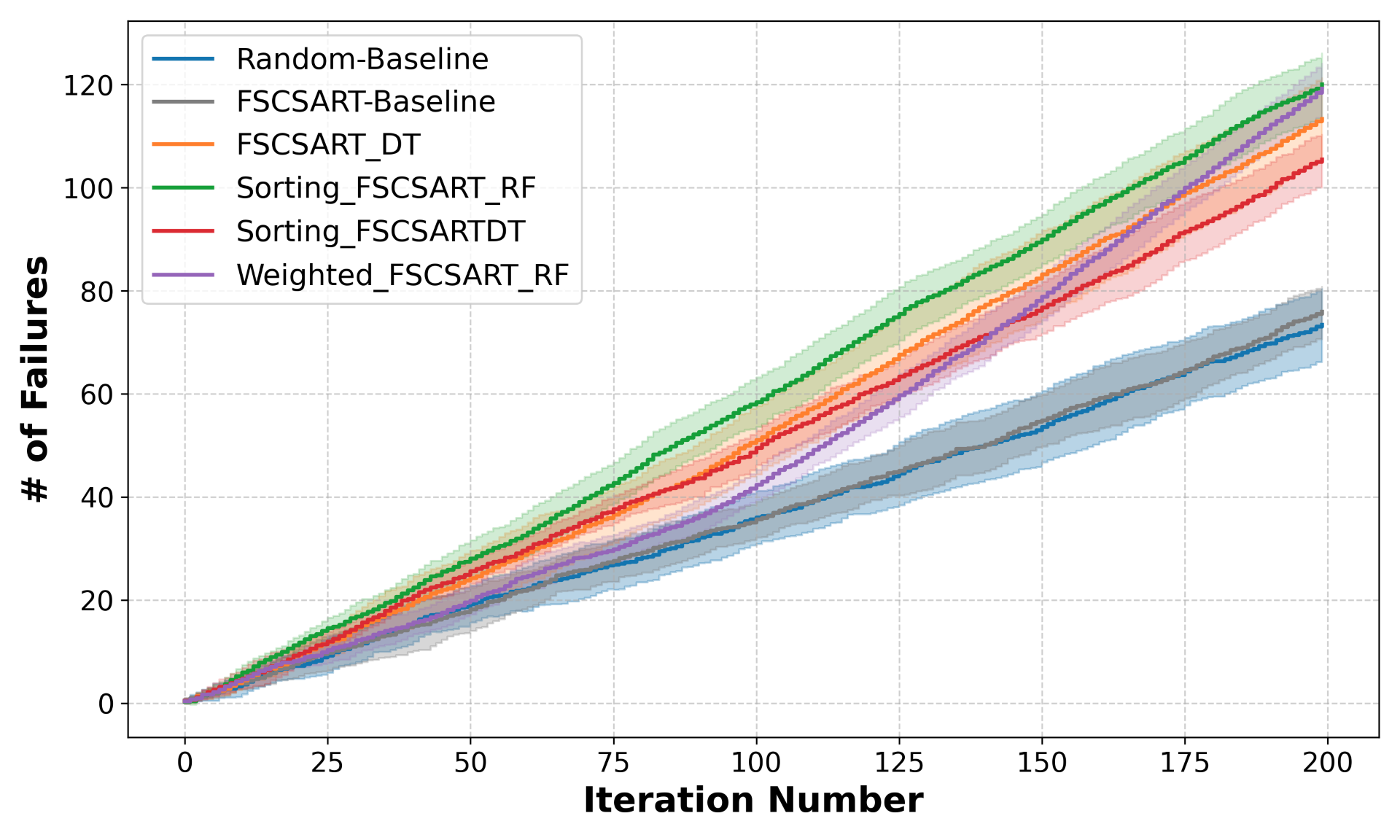}
        \caption{EO1}
        \label{fig:cum_eo1}
    \end{subfigure}
    \hfill
    \begin{subfigure}[b]{0.48\linewidth}
        \centering
        \includegraphics[width=\linewidth]{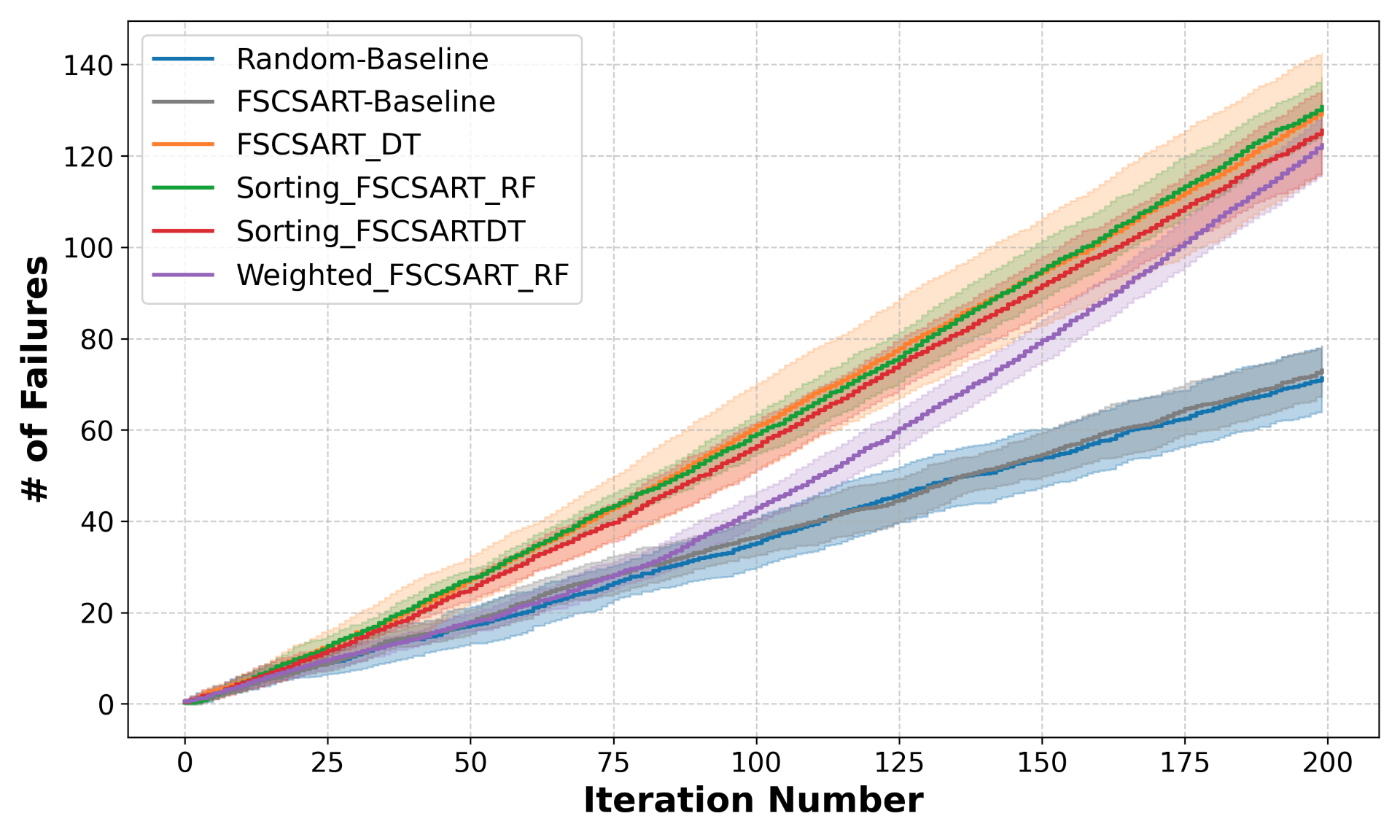}
        \caption{GR00T-N1.6}
        \label{fig:cum_gr00t}
    \end{subfigure}

    \vspace{0.3em}

    \begin{subfigure}[b]{0.48\linewidth}
        \centering
        \includegraphics[width=\linewidth]{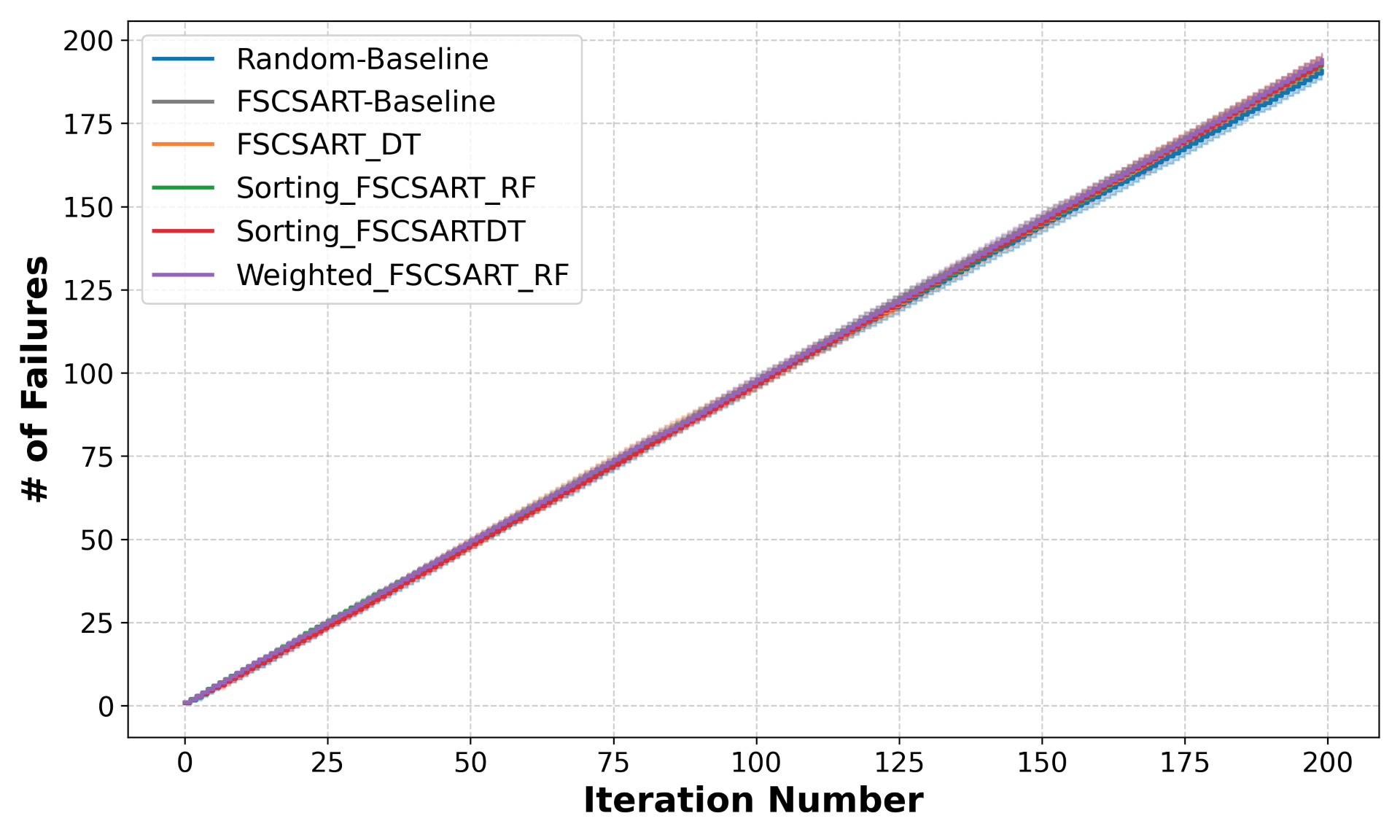}
        \caption{OpenVLA-7b}
        \label{fig:cum_openvla}
    \end{subfigure}
    \hfill
    \begin{subfigure}[b]{0.48\linewidth}
        \centering
        \includegraphics[width=\linewidth]{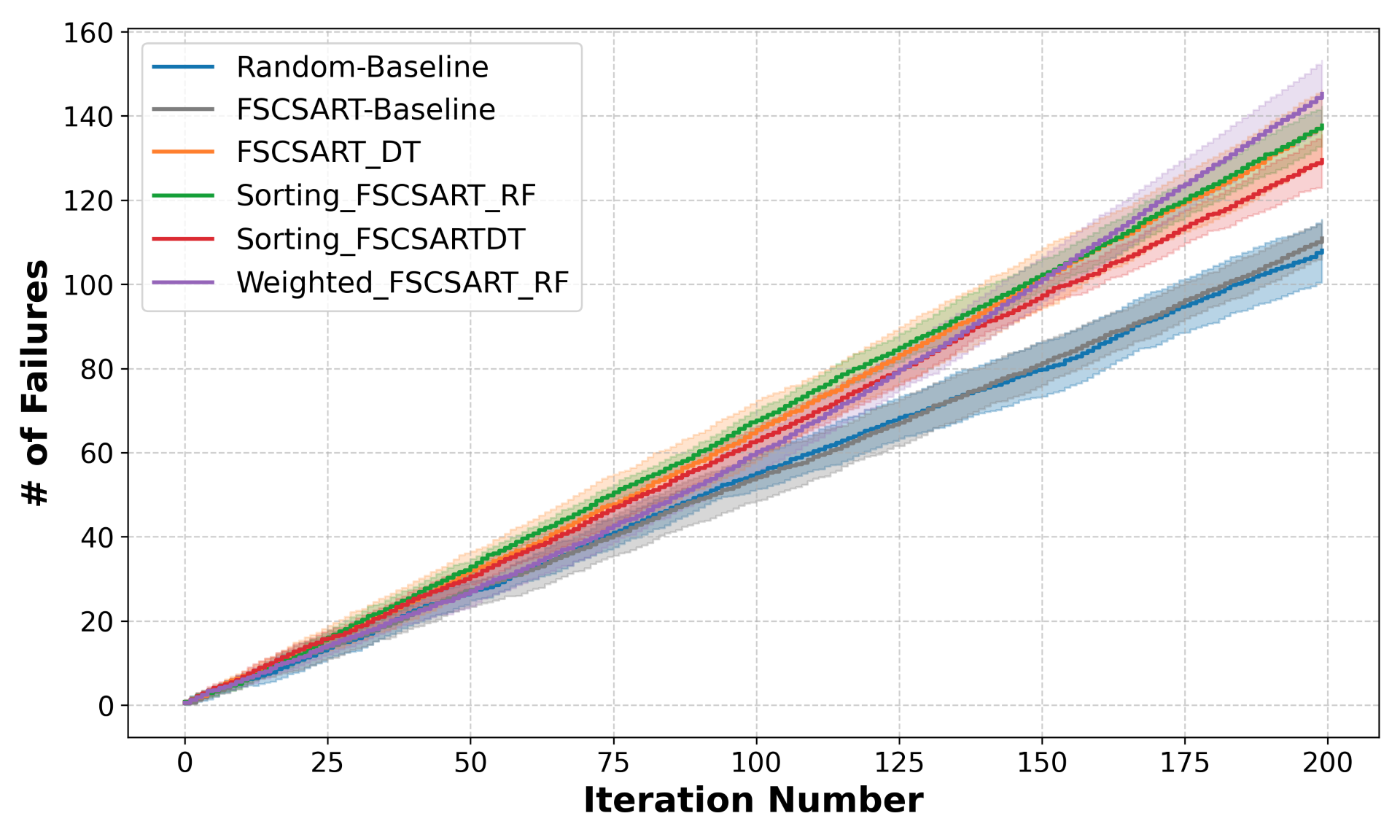}
        \caption{\pimodel}
        \label{fig:cum_pi0}
    \end{subfigure}

    \caption{Cumulative failures discovered during 200 testing iterations across four VLA models. Each curve represents one testing method, while shaded regions show the interquartile range over 10 runs. The proposed methods consistently discover failures faster than both baselines on EO1, GR00T-N1.6, and \pimodel, whereas all methods converge on OpenVLA-7b due to a ceiling effect.}
    \label{fig:cumulative_failures1}
\end{figure}

RQ1 evaluates whether the proposed ML-guided algorithms improve failure discovery beyond both baseline approaches: B1 (Random) and B2 (FSCSART). Across EO1, GR00T-N1.6, and \pimodel, the proposed methods consistently achieved substantially higher FR, with gains reaching $+29.7$~\% over Random and $+28.8$~\% over FSCSART on GR00T-N1.6. Similar improvements are observed for EO1 and \pimodel, where the RF-guided variants achieved the strongest FR performance. In contrast, OpenVLA-7b represents a ceiling case where all approaches already achieved very high FR values (95--98\% ), indicating that failures are trivially discoverable for this model. Importantly, the FR gains of the proposed methods were achieved without sacrificing exploration quality. TC, TCF, and FOC remained comparable to or higher than both baselines across most VLAs, demonstrating that the proposed methods maintain broad behavioral and object-level exploration while discovering more failures. The RF-based variants, particularly \texttt{Sorting\_RF} and \texttt{Weighted\_RF}, provided the most balanced performance by simultaneously improving FR, TCF, and FOC. In comparison, DT-guided approaches occasionally over-focused on dominant failure regions, slightly reducing diversity metrics despite strong FR gains. Another important observation is that B1 and B2 achieved nearly identical FR values on all non-ceiling VLAs, showing that diversity-driven ART alone contributes little improvement over random testing. This confirms that the primary source of performance improvement comes from the ML guidance component rather than the ART diversity mechanism itself. Figure~\ref{fig:cumulative_failures1} further illustrates these trends through cumulative failure discovery curves, where the proposed methods separate early from both baselines and maintain their advantage throughout the testing process. Detailed metric-wise analysis, cumulative failure curves, and box-plot distributions are provided in Appendix~\ref{sec:appendix_rq1_details} and Appendix~\ref{sec:appendix_boxplots}.
\noindent\fbox{%
\parbox{\dimexpr\linewidth-2\fboxsep\relax}{%
\textbf{Answer for RQ1.}
Our approaches significantly outperform both baselines in failure detection, with gains of up to $+29.7$~\% in FR. The similar performance of B1 and B2 suggests that ART diversity alone provides limited benefit. RF-based variants also maintain strong TC, TCF, and FOC, achieving better failure discovery without reducing exploration diversity.}}
\vspace{-1.5em}
\subsection{RQ2: Impact of the ML Model (DT vs.\ RF) on Failure Detection}
\label{sec:rq2}
RQ2 investigates the contribution of the ML guidance component and compares the effectiveness of DT and RF models for failure-aware test generation. The results show that incorporating ML guidance into FSCS-ART substantially improves failure discovery compared with diversity-only ART. In particular, \texttt{FSCSART\_DT} consistently achieved large FR gains over the original FSCSART baseline, confirming that learned failure prediction was the primary driver of performance improvement. Comparing DT- and RF-based approaches further reveals that RF-guided variants provided a better balance between exploitation and exploration. While DT guidance can strongly increase FR, it occasionally over-focused on dominant failure regions, reducing behavioral diversity metrics such as TCF. In contrast, RF-based methods, particularly \texttt{Sorting\_RF} and \texttt{Weighted\_RF}, maintained higher TCF and FOC while achieving equal or higher FR across most VLAs. This behavior might be because RF models generate smoother and more robust failure-probability estimates through ensemble learning, avoiding the hard partitioning effects of a single DT. Among all methods, \texttt{Weighted\_RF} provided the most consistent overall performance by combining RF-based failure prediction with diversity-aware candidate selection. These findings demonstrate that the choice of the ML model significantly influences both the quantity and diversity of discovered failures, with RF-based guidance offering the most effective trade-off. Detailed quantitative comparisons are provided in Appendix~\ref{sec:appendix_rq2_details}.
\vspace{-1mm}
\vspace{0.5em}
\noindent\fbox{%
\parbox{\dimexpr\linewidth-2\fboxsep\relax}{%
\textbf{Answer for RQ2.}
Incorporating ML guidance into FSCS-ART significantly improves failure detection compared with diversity-only ART. Although DT-based guidance achieves strong FR gains, RF-based variants provide a better balance between failure discovery and diversity preservation, with \texttt{Weighted\_RF} delivering the most consistent overall performance.}}
\vspace{-2.5mm}

\vspace{-2mm}
\section{Limitations}
\label{sec:limitations}
\vspace{-3mm}

While our approach shows strong gains, several limitations remain. One relates to \textbf{Task specificity.} We evaluate a single manipulation task, which may limit generalizability to more complex or long-horizon settings. However, failures exposed in this comparatively simple, high-performing task suggest that similar or stronger weaknesses may arise in harder scenarios. Future work will extend the evaluation to multi-task benchmarks such as CALVIN, LIBERO, and ManiSkill2. Another limitation refers to \textbf{object diversity}. Our experiments used a restricted pool of only seven objects, whereas real deployments involve much broader variation in shape, texture, and affordances. We will address this by incorporating larger object sets and procedural scene/object generation in extended versions of this paper. A third limitation refers to \textbf{simulation-based evaluation.} Our study is conducted in simulation because large-scale real-robot failure-seeking experiments across multiple models and repeated scenes are currently impractical due to cost, reset effort, safety risks, and hardware constraints. We therefore use simulation as a controlled preliminary validation step, following common practice in robot learning. The last limitation relates to \textbf{stochasticity.}
Both VLA execution and test generation are stochastic. To reduce sensitivity to individual seeds, each configuration is repeated 10 times and results are reported using aggregate statistics across runs.\vspace{-1 em}
\section{Conclusion}
\label{sec:conclusion}
\vspace{-2mm}
VLA evaluation still relies largely on static benchmarks, which provide only a limited view of robustness in high-dimensional embodied spaces where failures are sparse, clustered, and configuration-dependent. In this paper, we reframed VLA evaluation as an active failure discovery problem and proposed a failure-aware test generation approach that combines diversity-driven exploration with machine-learning-guided failure prediction. Across four state-of-the-art VLA models, our approaches substantially outperformed random testing and diversity-driven ART baselines, increasing FR by up to 29.7 \% over random testing and 28.8 points over FSCS-ART. These results show that unguided exploration is insufficient and that VLA failure regions exhibit learnable structure that can be exploited to generate more effective and diverse test suites. Overall, our findings provide evidence that the community should move beyond passive static benchmarking toward adaptive, failure-seeking evaluation pipelines that better expose systematic weaknesses before deployment.

\section*{Acknowledgments}
This work is supported by the InnoGuard Marie Skłodowska-Curie Doctoral Network (Grant Agreement No. 101169233). 
Arusa Kanwal, Pablo Valle, Aitor Arrieta are part of the Software and Systems Engineering research group of Mondragon Unibertsitatea (IT1519-22), supported by the Department of Education, Universities and Research of the Basque Country. Pablo Valle is supported by the Pre-doctoral Program for the Formation of Non-Doctoral Research Staff of the Education Department of the Basque Government (Grant n. PRE\_2025\_2\_0252).

\bibliographystyle{unsrt}  
\bibliography{references}

\appendix
\input{appendix}

\end{document}

%% file: appendix.tex

\clearpage

\section{RQ1: Performance Against Both Baselines}
\label{sec:appendix_rq1_details}
\label{sec:rq1}

\begin{figure}[!h]
    \centering
    \begin{subfigure}[b]{0.45\linewidth}
        \centering
        \includegraphics[width=\linewidth]{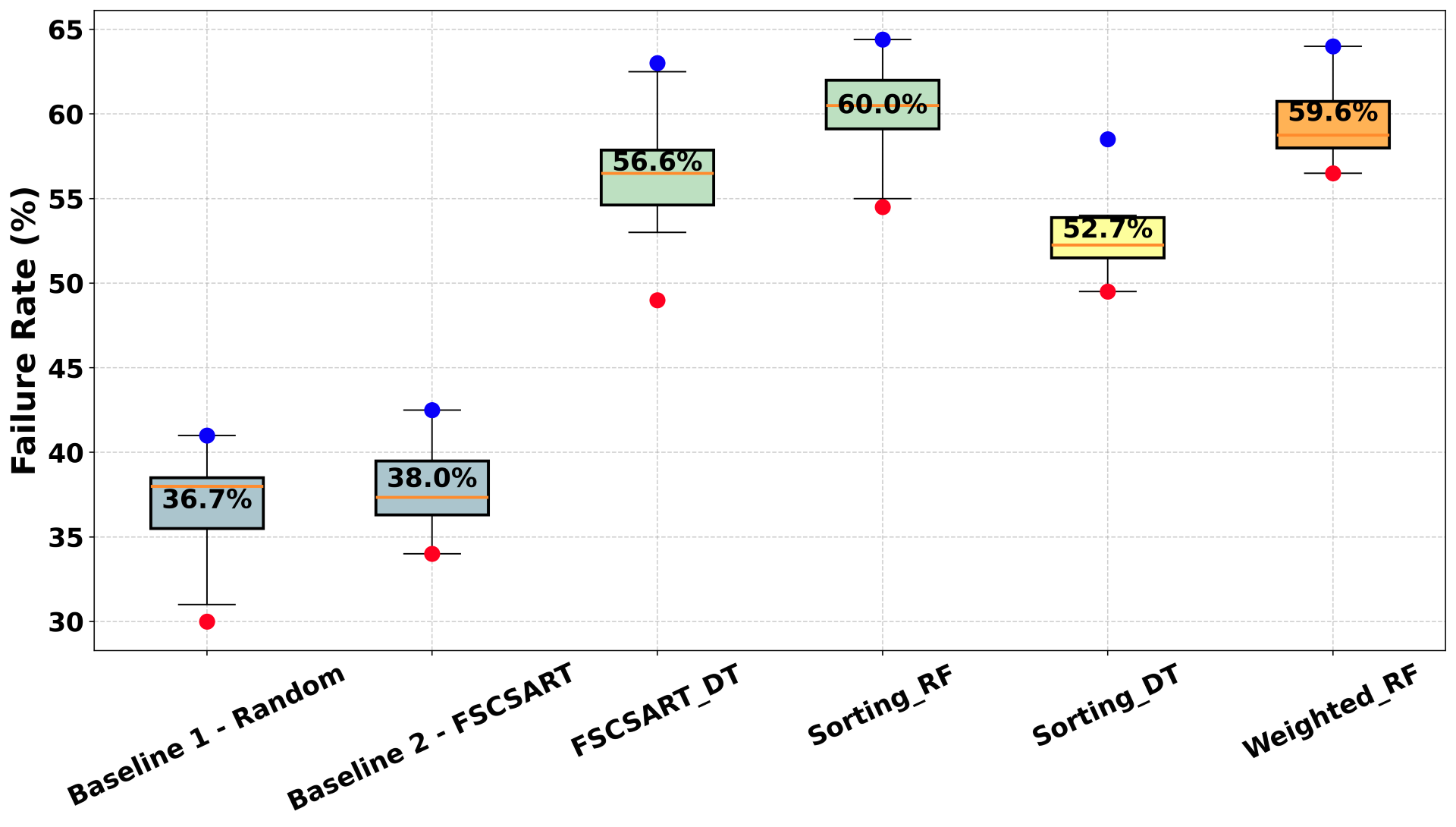}
        \caption{EO1}
        \label{fig:fr_eo1}
    \end{subfigure}
    \hfill
    \begin{subfigure}[b]{0.45\linewidth}
        \centering
        \includegraphics[width=\linewidth]{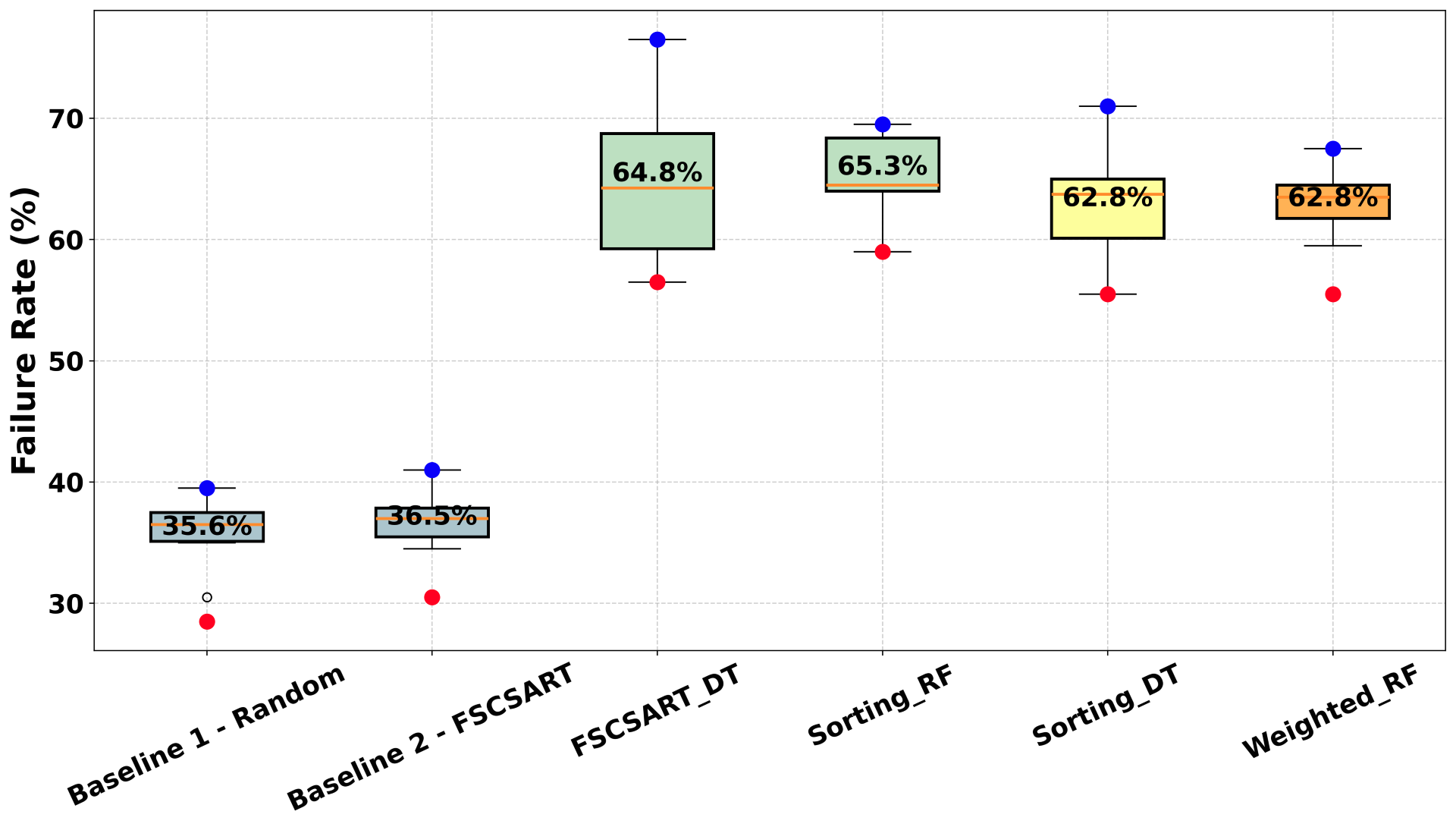}
        \caption{GR00T-N1.6}
        \label{fig:fr_gr00t}
    \end{subfigure}
    \vspace{0.0em}
    \begin{subfigure}[b]{0.45\linewidth}
        \centering
        \includegraphics[width=\linewidth]{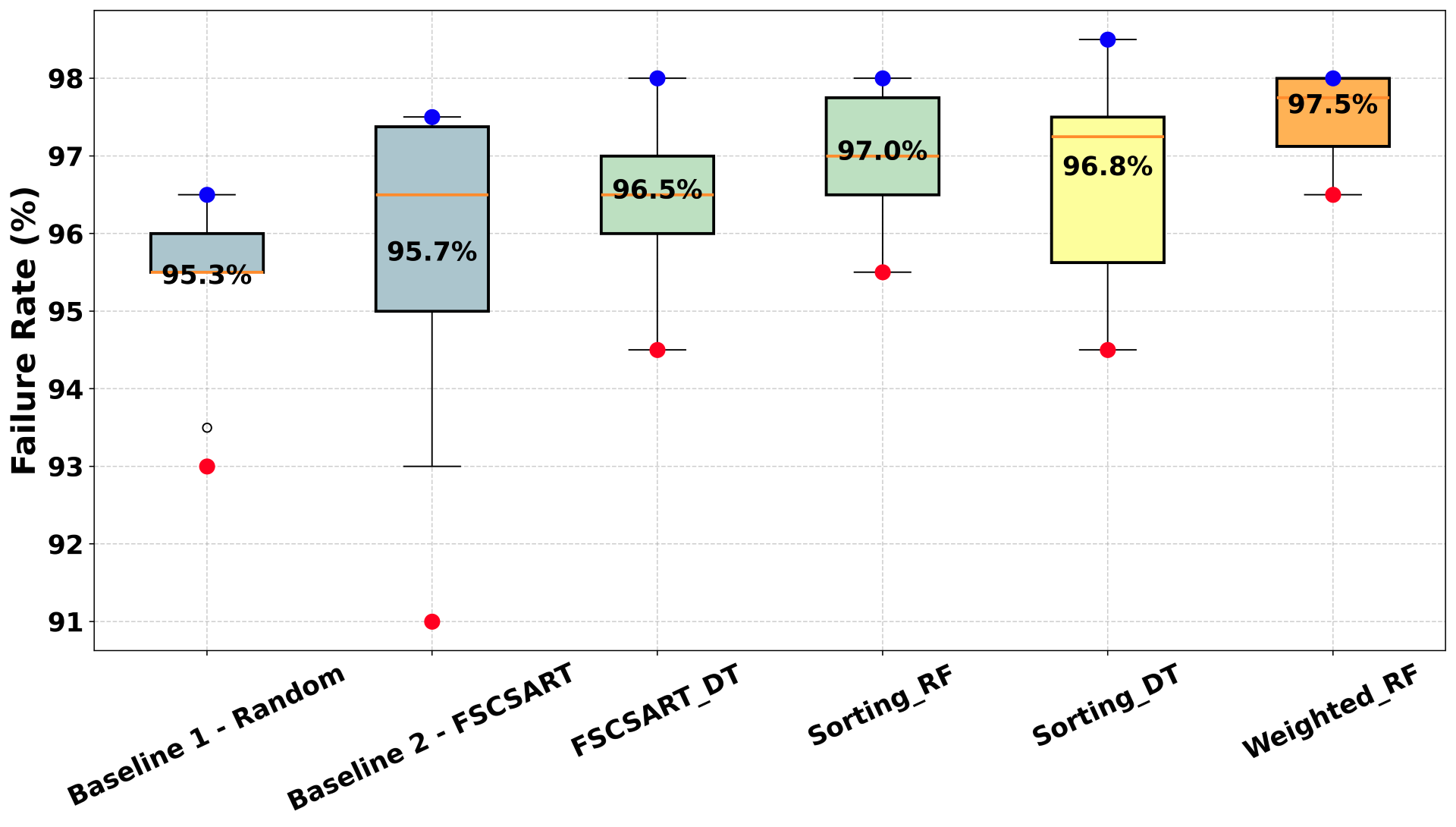}
        \caption{OpenVLA-7b}
        \label{fig:fr_openvla}
    \end{subfigure}
    \hfill
    \begin{subfigure}[b]{0.45\linewidth}
        \centering
        \includegraphics[width=\linewidth]{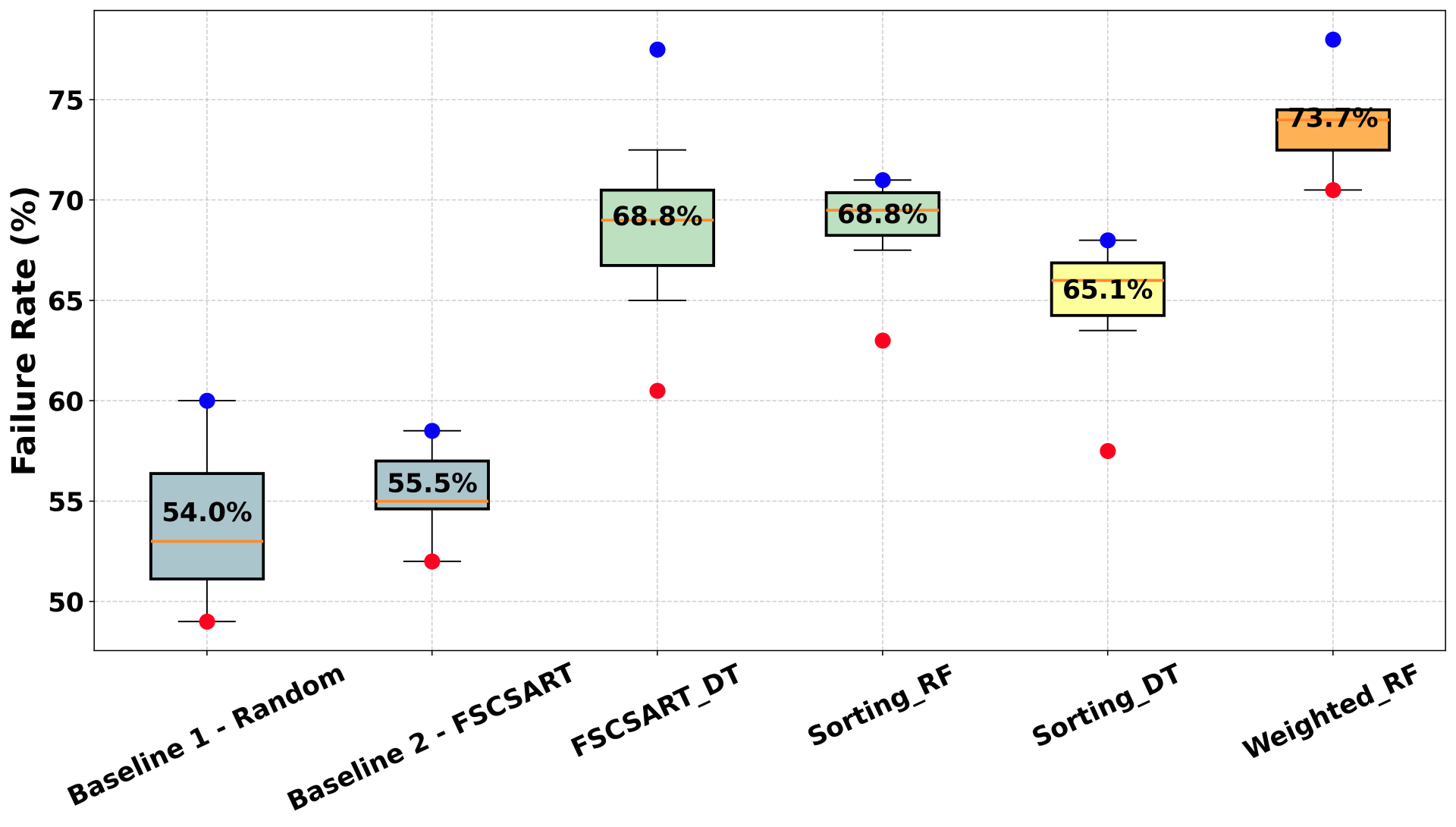}
        \caption{\pimodel}
        \label{fig:fr_pi0}
    \end{subfigure}
    \caption{Final Failure Rate (\%) distributions for all algorithms across four VLA models. Annotated values are medians.
    Proposed methods substantially exceed both B1:~Random and B2:~FSCSART on EO1, GR00T-N1.6, and \pimodel.}
    \label{fig:fr_boxplots}
\end{figure}

RQ1 serves as a sanity check for both our proposed ML-guided algorithms and the complexity of the underlying failure-discovery problem. B1 (Random) represents a minimum-effort testing strategy, while B2 (FSCSART) represents a strong diversity-driven strategy without learned failure guidance. If these baselines were sufficient, failure discovery would be largely reducible to random or diversity-only exploration. Instead, improvements over both baselines would indicate that VLA failures are structured and non-trivial to uncover, motivating the need for more complex test generation approaches.
\paragraph{Failure Rate (FR):}
Figure~\ref{fig:cumulative_failures1} shows the cumulative number of failures discovered over 200 test iterations. Detailed FR statistics are provided in Appendix~\ref{sec:appendix_tables}, Table~\ref{tab:fr}. The corresponding distributions are shown as box plots in Figure~\ref{fig:fr_boxplots} (Appendix~\ref{sec:appendix_boxplots}). From these results, we extract three key findings:

\textbf{(i) The proposed methods outperform both baselines.}
On EO1, the best proposed method (\texttt{Sorting\_RF}, 60.0\%) exceeds B1 by \textbf{+23.3~pp} and B2 by \textbf{+22.0~pp}.
On GR00T-N1.6, \texttt{Sorting\_RF} (65.3\%) outperforms B1 by \textbf{+29.7~pp} and B2 by \textbf{+28.8~pp}, nearly doubling both baselines' FR.
On \pimodel, \texttt{Weighted\_RF} (73.7\%) improves over B1 by \textbf{+19.7~pp} and over B2 by \textbf{+18.2~pp}.
These improvements are large, consistent across three architecturally distinct VLAs, and statistically robust over 10 repetitions.

\textbf{(ii) B1 and B2 are close to each other, but both are far below the proposed algorithms.}
On every non-ceiling VLA, B1 (Random) and B2 (FSCSART) differ by only 0.9–1.5~pp in FR, confirming that diversity-driven ART alone without a learned failure model provides little benefit over random sampling in terms of failure detection rate. The large gap between B2 and the proposed methods (up to $+28.8$~pp) therefore directly attributes the improvement to the ML guidance component, not to the ART diversity mechanism.

\textbf{(iii) OpenVLA-7b is a ceiling case.}
All six algorithms cluster between 95.3\% and 97.5\% on OpenVLA-7b, reflecting near-universal model susceptibility to the generated scenes. OpenVLA is therefore considered a very weak VLA, and there is no justification to use complex testing methods for it, unlike the remaining VLAs. Even here, the proposed methods modestly exceed both baselines ($+0.8$ to $+2.2$~pp over B1 and $+0.4$ to $+1.8$~pp over B2).

The cumulative curves in Figure~\ref{fig:cumulative_failures1} reinforce these findings. On EO1, GR00T-N1.6, and \pimodel, the three tiers are visually clear: B1 (lowest), B2 (middle, marginally above B1), and the cluster of proposed methods (highest), separating from both baselines within the first 50 iterations and maintaining that advantage for the remaining 150 iterations.

\begin{figure}[!h]
    \centering
    \begin{subfigure}[b]{0.45\linewidth}
        \centering
        \includegraphics[width=\linewidth]{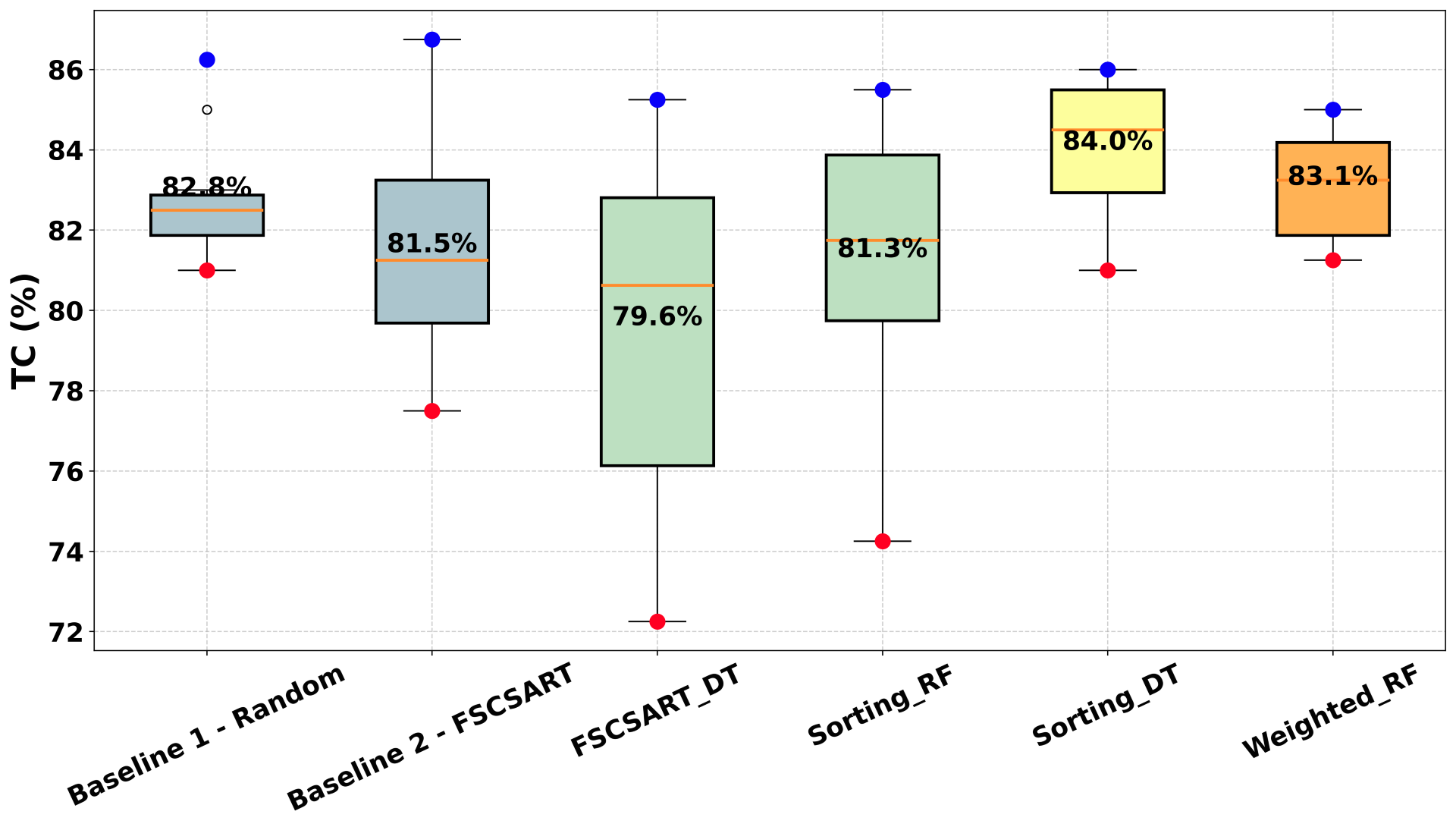}
        \caption{EO1}
        \label{fig:tc_eo1}
    \end{subfigure}
    \hfill
    \begin{subfigure}[b]{0.45\linewidth}
        \centering
        \includegraphics[width=\linewidth]{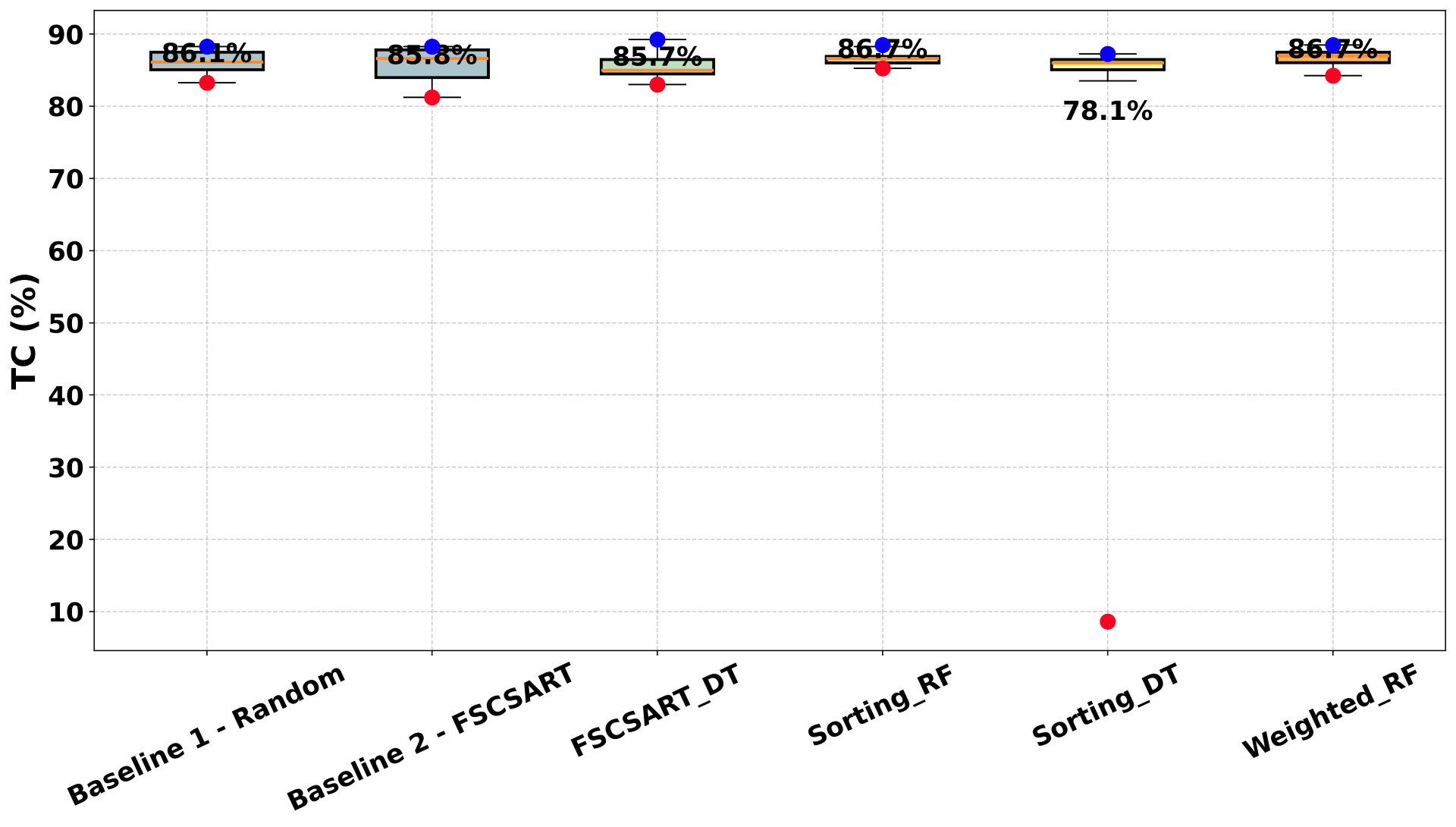}
        \caption{GR00T-N1.6}
        \label{fig:tc_gr00t}
    \end{subfigure}
    \vspace{0.0em}
    \begin{subfigure}[b]{0.45\linewidth}
        \centering
        \includegraphics[width=\linewidth]{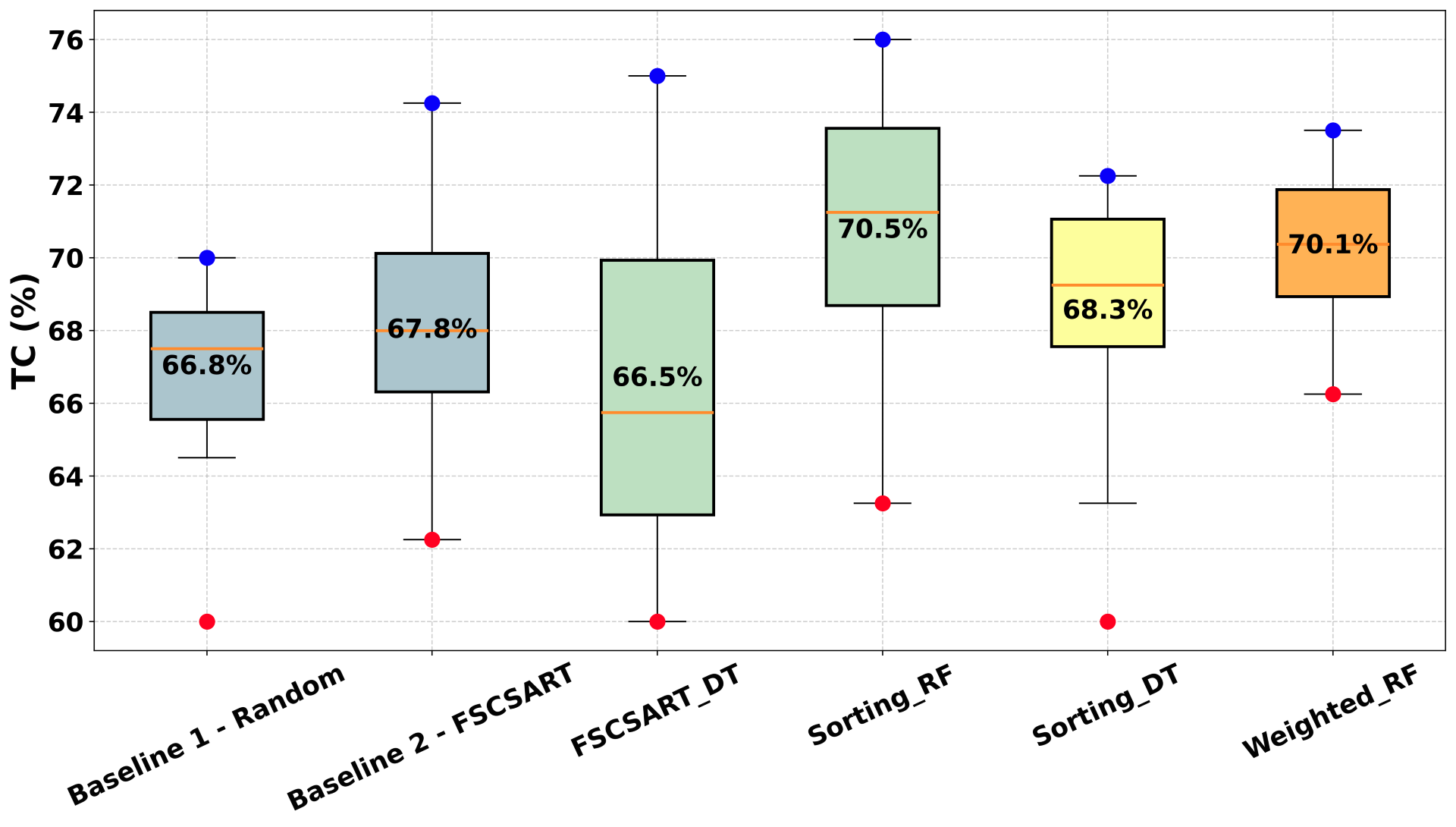}
        \caption{OpenVLA-7b}
        \label{fig:tc_openvla}
    \end{subfigure}
    \hfill
    \begin{subfigure}[b]{0.45\linewidth}
        \centering
        \includegraphics[width=\linewidth]{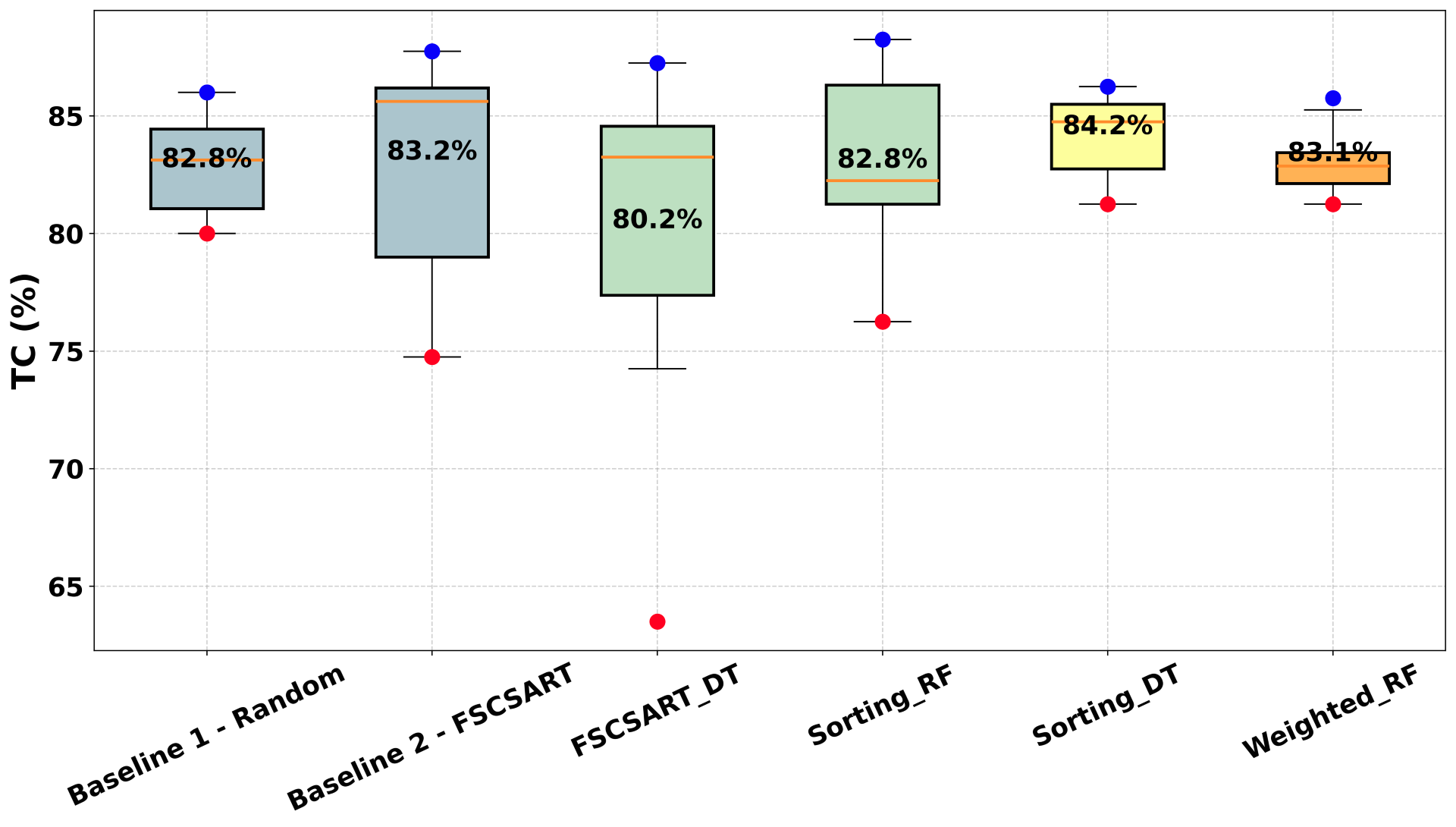}
        \caption{\pimodel}
        \label{fig:tc_pi0}
    \end{subfigure}
    \caption{Trajectory Coverage (\%) distributions. Proposed methods are broadly competitive with or exceed both baselines, confirming that ML-guided failure exploitation does not collapse behavioural exploration.}
    \label{fig:tc_boxplots}
\end{figure}

\begin{figure}[!h]
    \centering
    \begin{subfigure}[b]{0.45\linewidth}
        \centering
        \includegraphics[width=\linewidth]{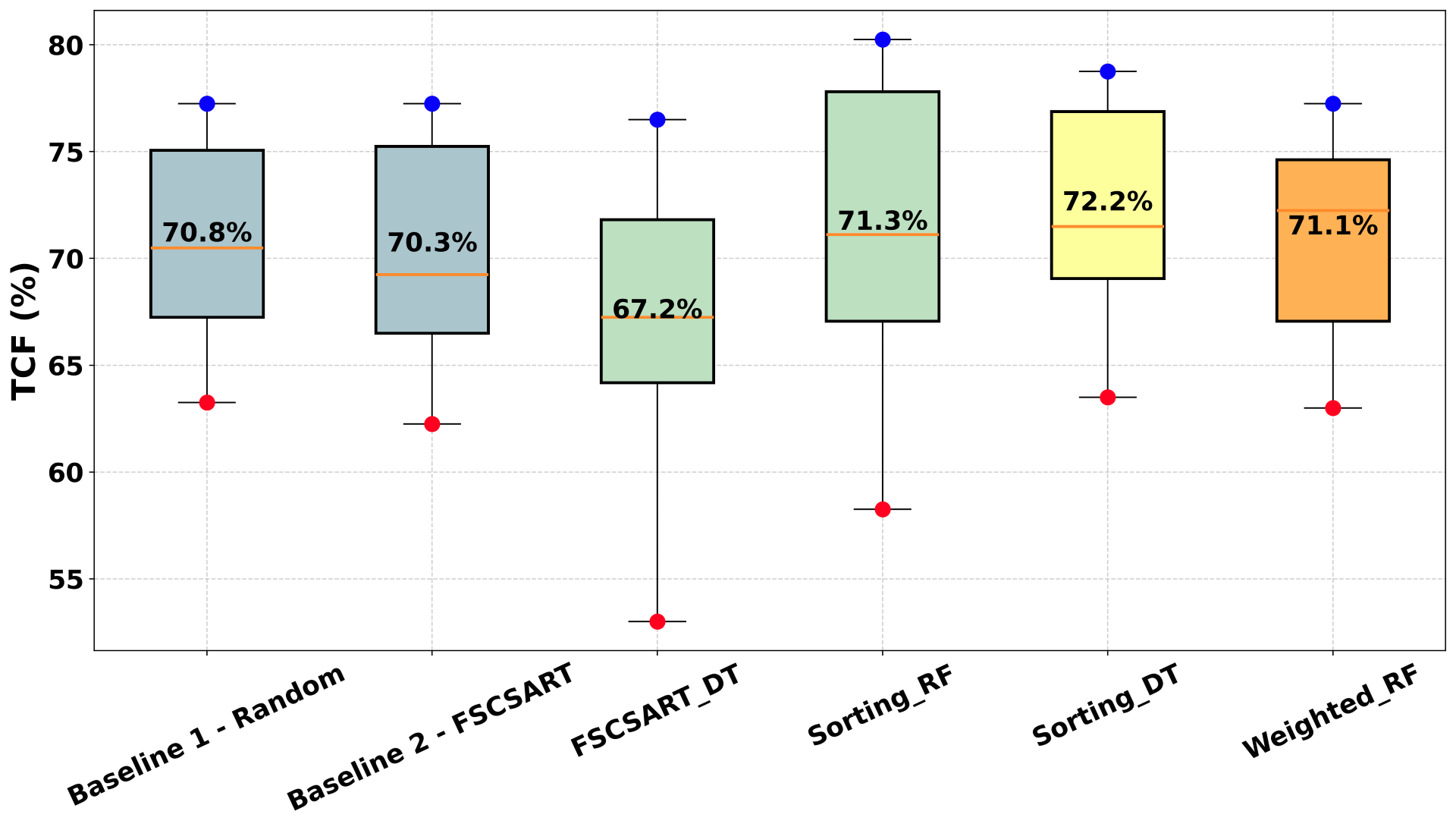}
        \caption{EO1}
        \label{fig:tcf_eo1}
    \end{subfigure}
    \hfill
    \begin{subfigure}[b]{0.45\linewidth}
        \centering
        \includegraphics[width=\linewidth]{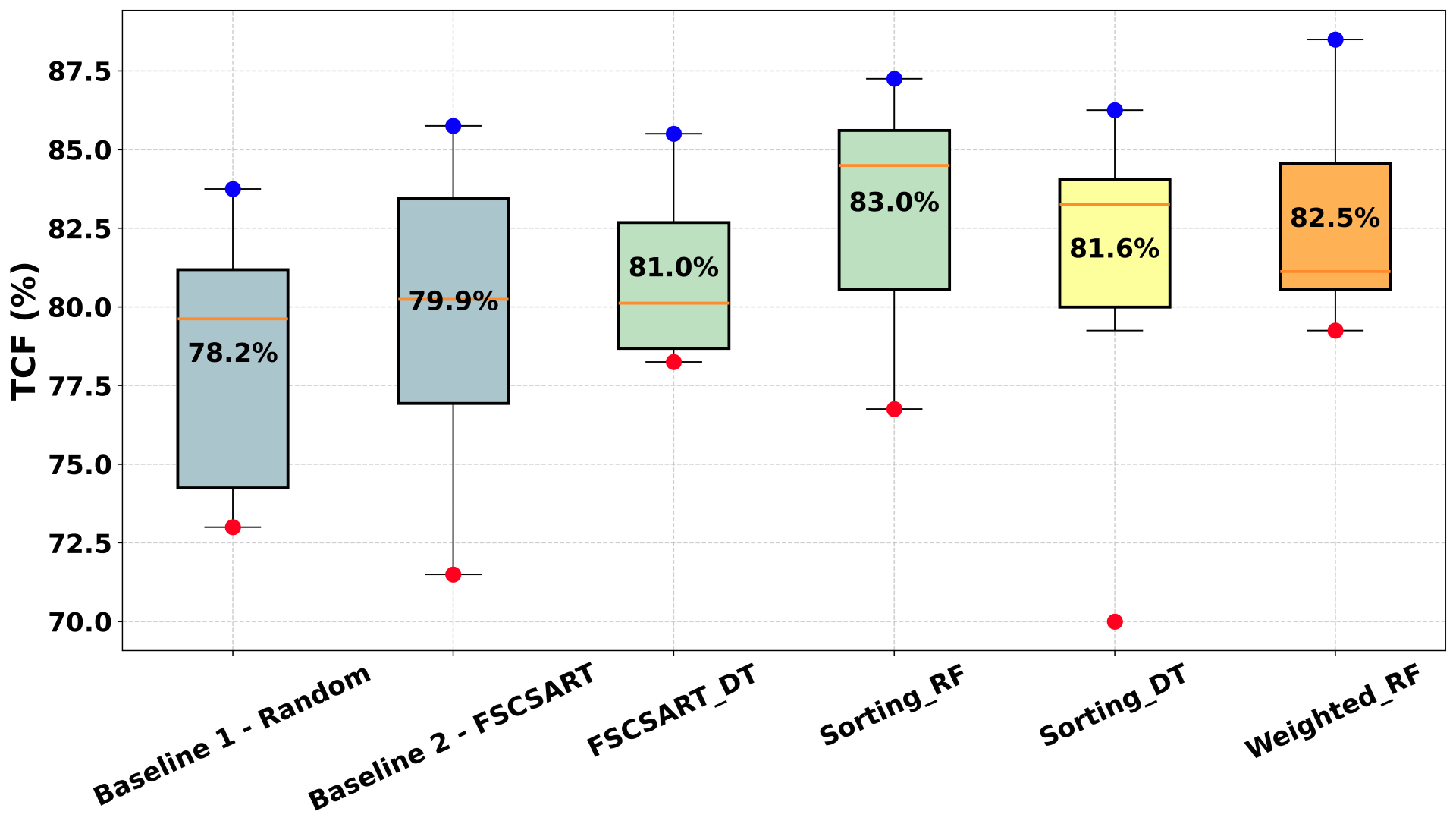}
        \caption{GR00T-N1.6}
        \label{fig:tcf_gr00t}
    \end{subfigure}
    \vspace{0.0em}
    \begin{subfigure}[b]{0.45\linewidth}
        \centering
        \includegraphics[width=\linewidth]{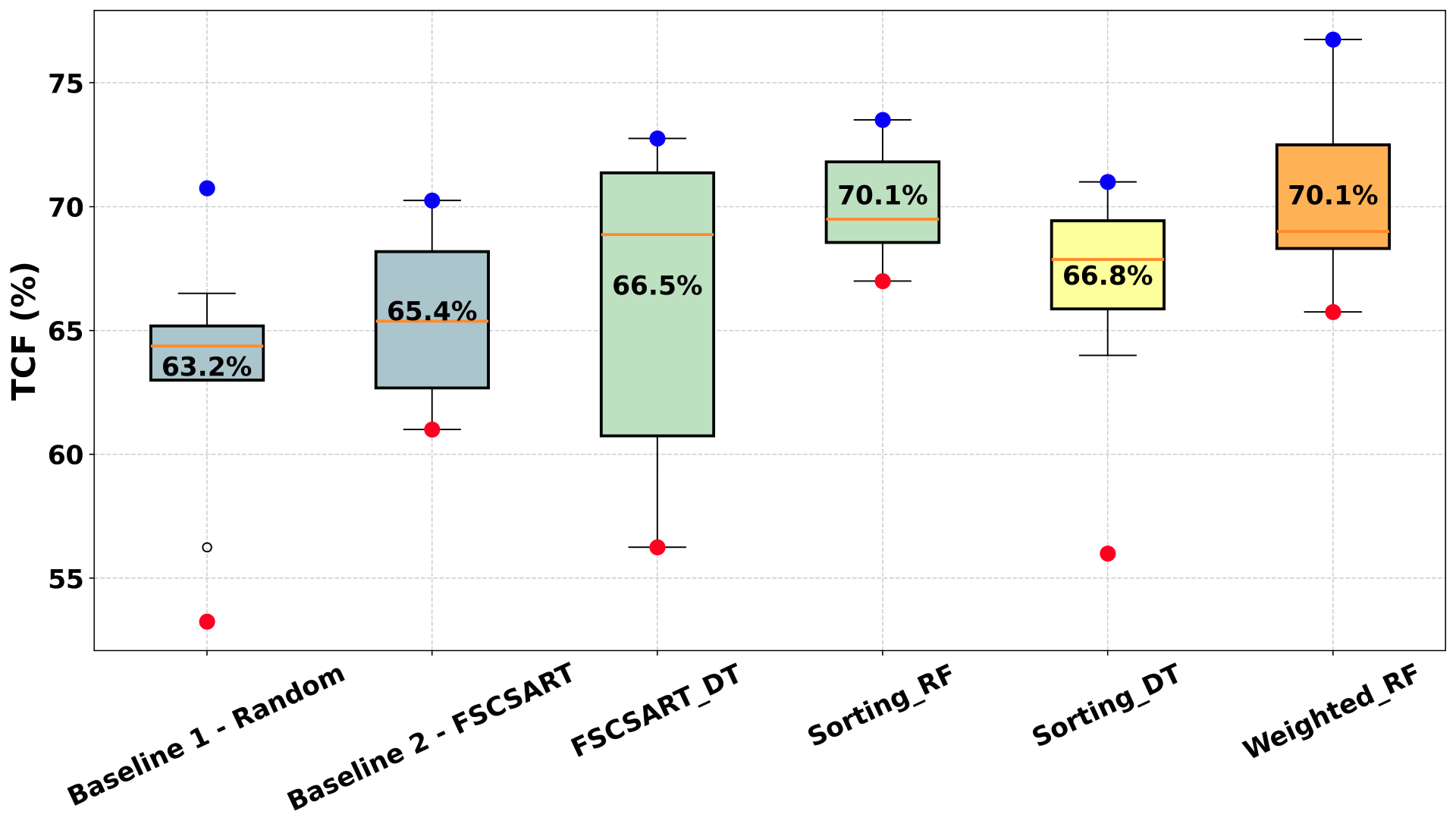}
        \caption{OpenVLA-7b}
        \label{fig:tcf_openvla}
    \end{subfigure}
    \hfill
    \begin{subfigure}[b]{0.45\linewidth}
        \centering
        \includegraphics[width=\linewidth]{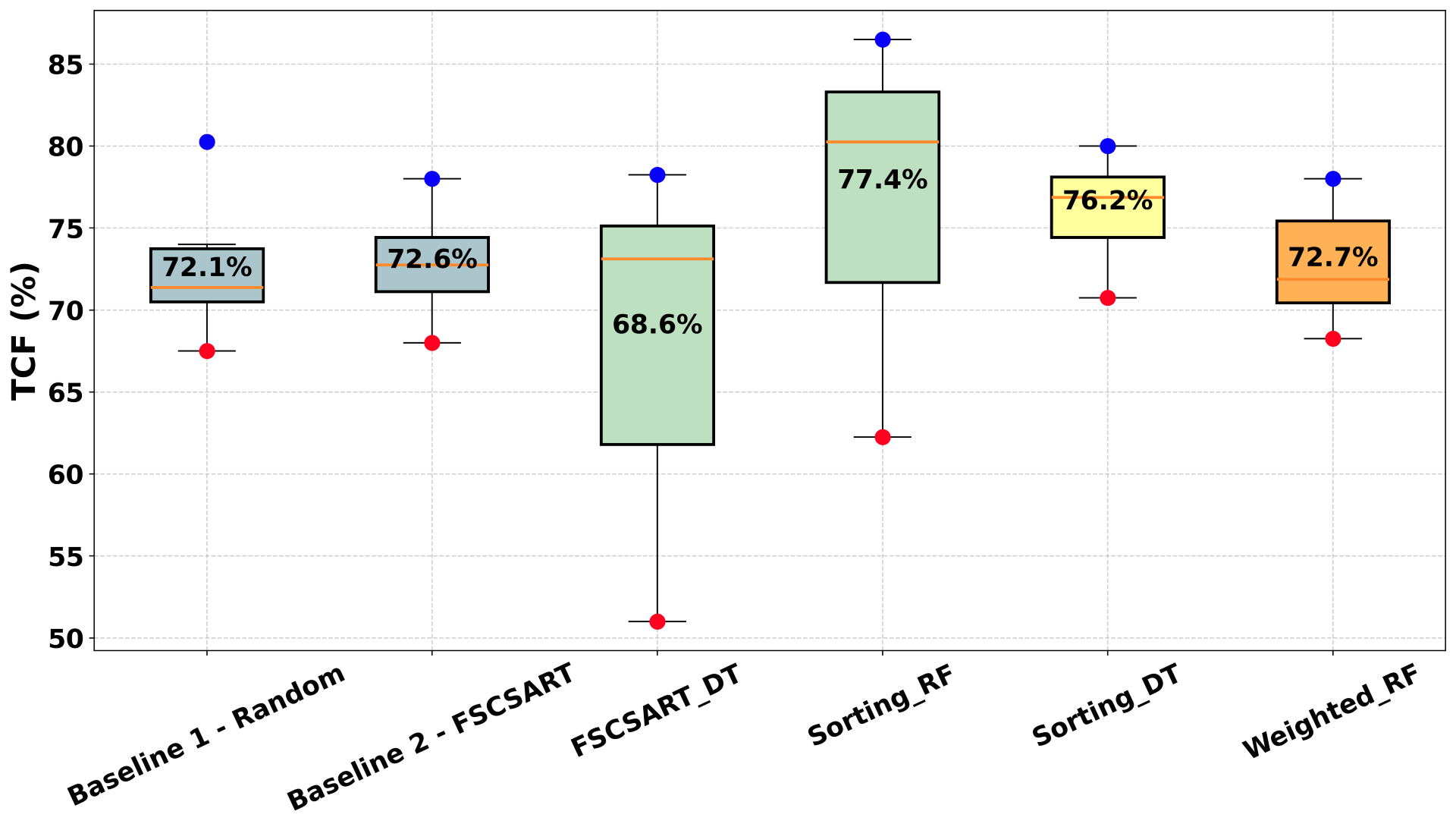}
        \caption{\pimodel}
        \label{fig:tcf_pi0}
    \end{subfigure}
    \caption{Trajectory Coverage under Failure, TCF (\%) distributions. RF-based variants (\texttt{Sorting\_RF}, \texttt{Weighted\_RF}) exceed both baselines on most VLAs, indicating that ensemble-guided generation discovers behaviourally diverse failures. The DT-only variant (\texttt{FSCSART\_DT}) occasionally falls below B2 due to over-exploitation of a single failure cluster.}
    \label{fig:tcf_boxplots}
\end{figure}

\begin{figure}[!ht]
    \centering
    \begin{subfigure}[b]{0.45\linewidth}
        \centering
        \includegraphics[width=\linewidth]{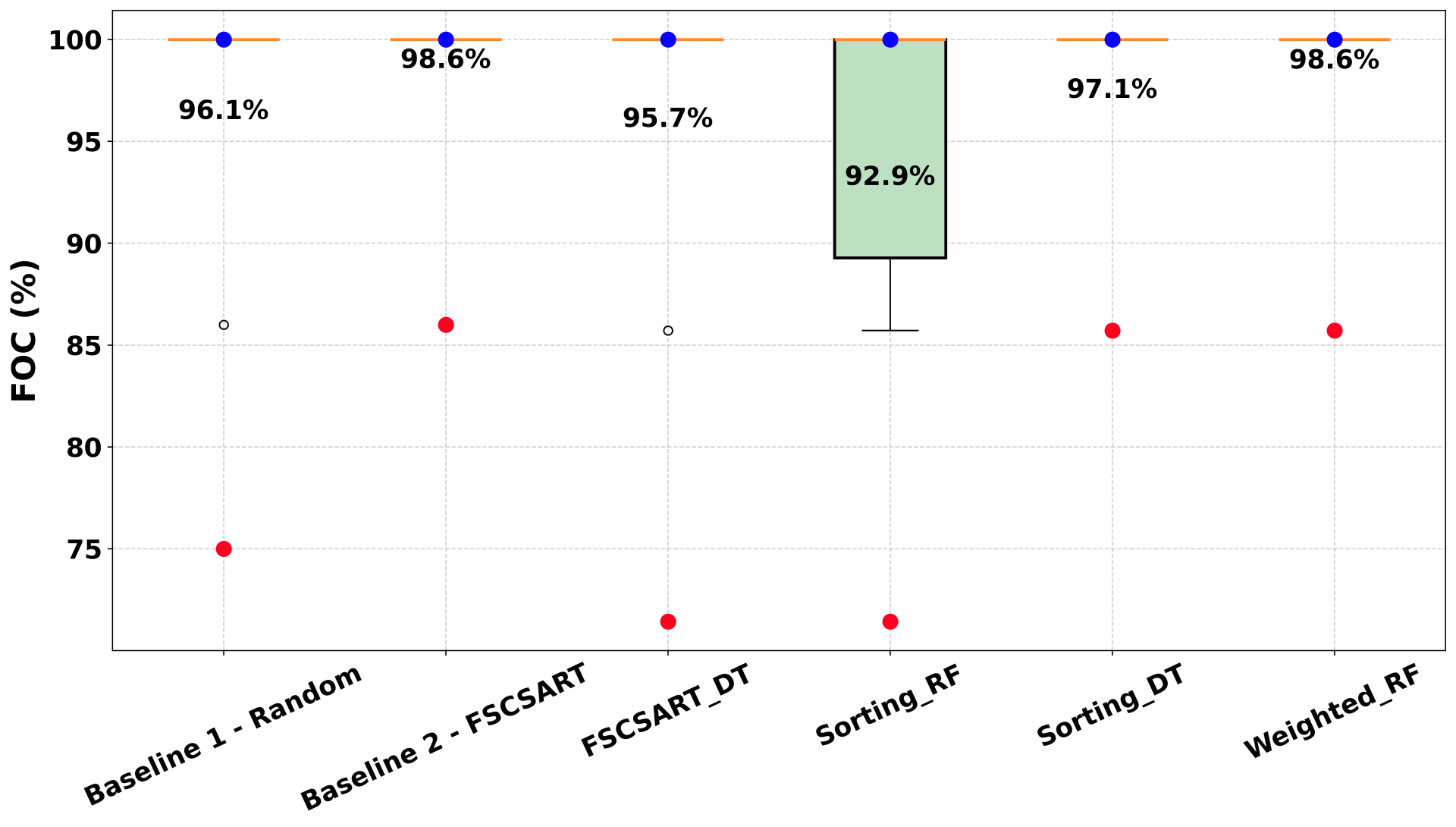}
        \caption{EO1}
        \label{fig:foc_eo1}
    \end{subfigure}
    \hfill
    \begin{subfigure}[b]{0.45\linewidth}
        \centering
        \includegraphics[width=\linewidth]{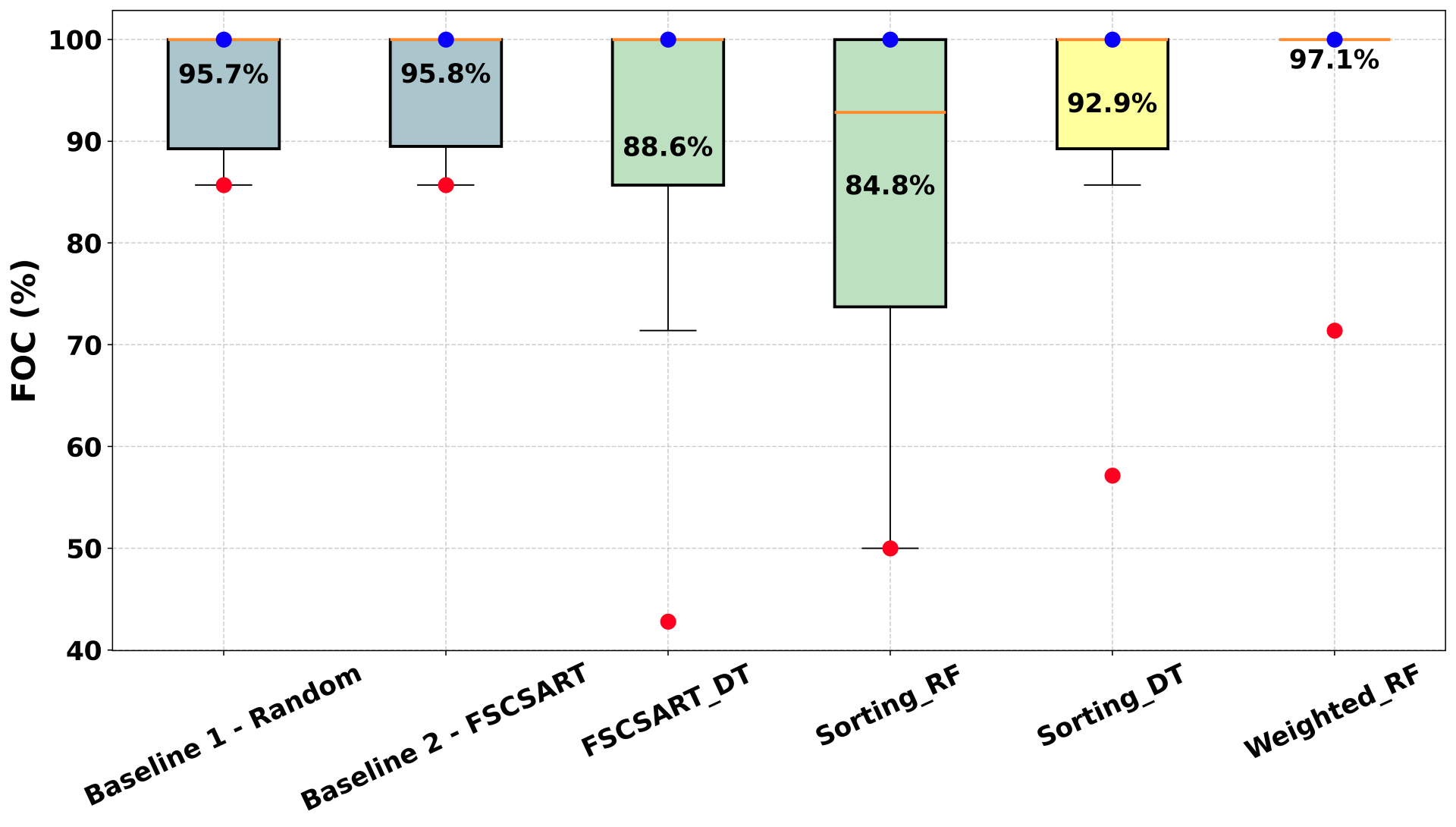}
        \caption{GR00T-N1.6}
        \label{fig:foc_gr00t}
    \end{subfigure}
    \vspace{0.0em}
    \begin{subfigure}[b]{0.45\linewidth}
        \centering
        \includegraphics[width=\linewidth]{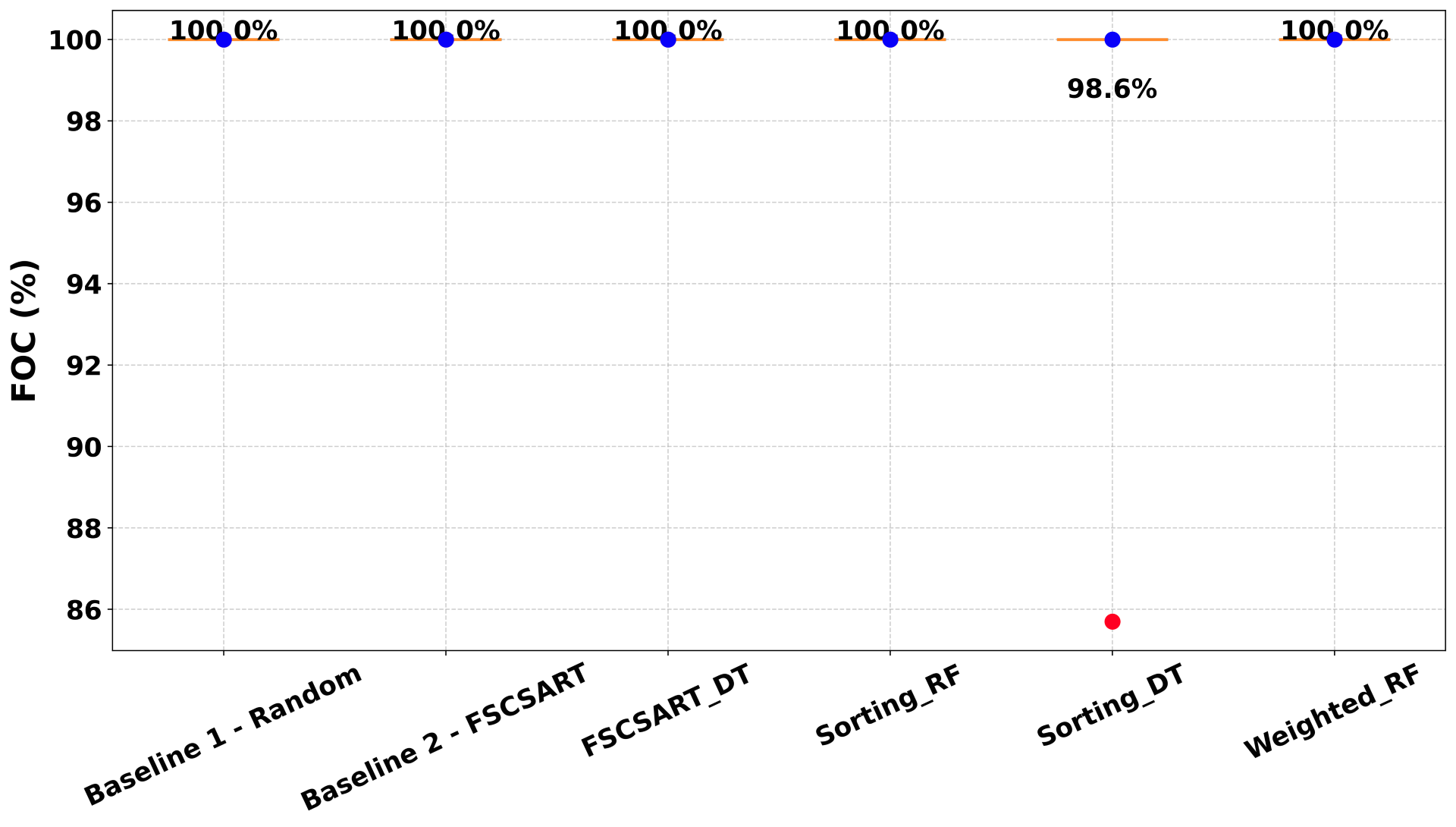}
        \caption{OpenVLA-7b}
        \label{fig:foc_openvla}
    \end{subfigure}
    \hfill
    \begin{subfigure}[b]{0.45\linewidth}
        \centering
        \includegraphics[width=\linewidth]{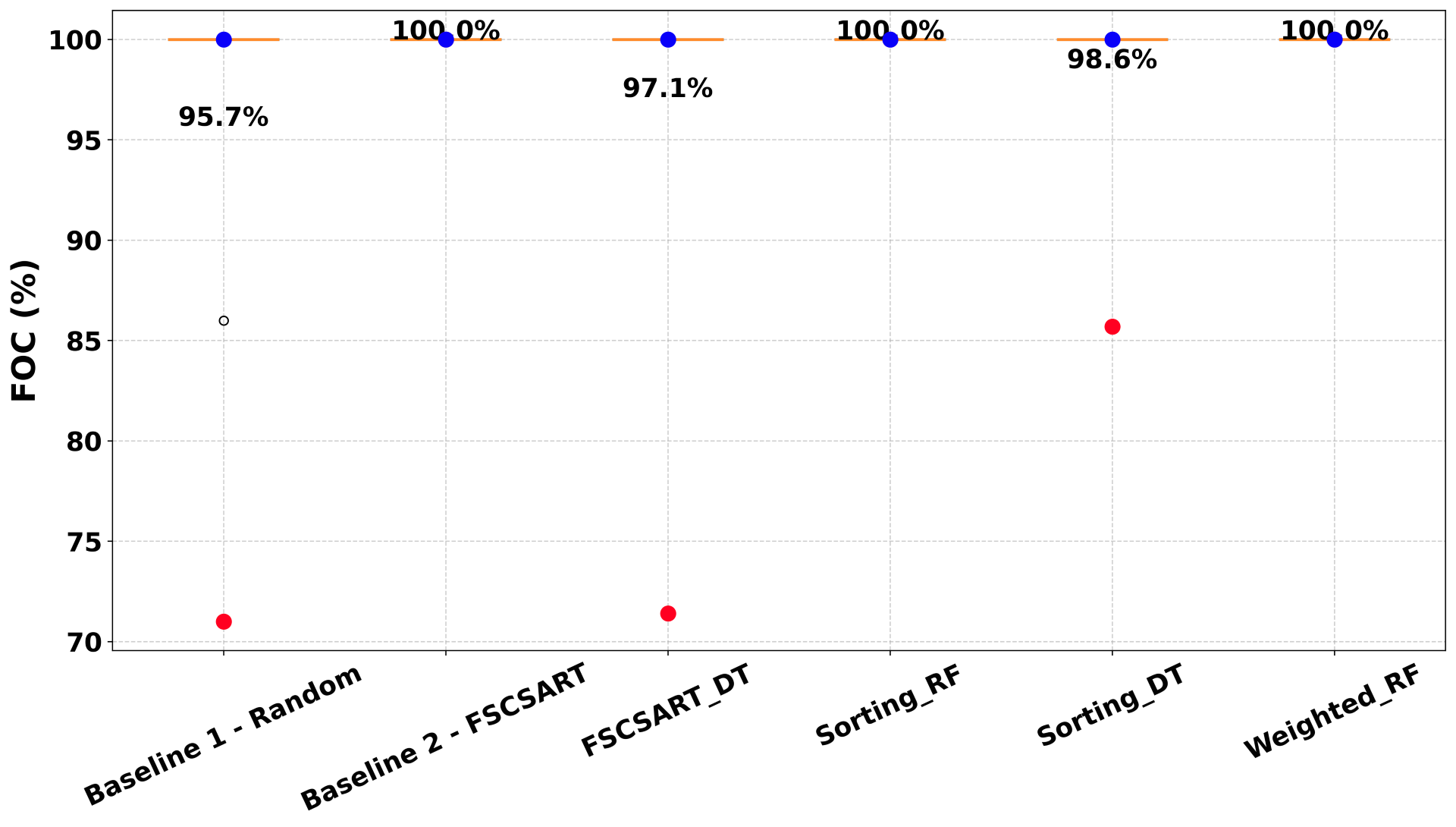}
        \caption{\pimodel}
        \label{fig:foc_pi0}
    \end{subfigure}
    \caption{Failed Object Coverage (\%) distributions. \texttt{Weighted\_RF} consistently achieves the highest FOC while simultaneously delivering the largest FR gains, demonstrating that RF-guided weighted scoring avoids object-concentration bias.}
    \label{fig:foc_boxplots}
\end{figure}

\paragraph{Trajectory Coverage (TC):}
Higher FR is only meaningful if behavioral diversity is not collapsed. Detailed TC statistics are reported in Appendix~\ref{sec:appendix_tables}, Table~\ref{tab:tc}, with the full distributions shown in Figure~\ref{fig:tc_boxplots} (Appendix~\ref{sec:appendix_boxplots}). The proposed methods do not sacrifice trajectory diversity and in several cases exceed the baselines. On EO1, \texttt{Sorting\_DT} surpasses both baselines (84.0\% vs.\ 82.8\% for B1 and 81.5\% for B2). On GR00T-N1.6, five of six methods converge to TC~$\approx$86\%, matching both baselines; only \texttt{Sorting\_DT} drops to 78.1\%, indicating a stronger exploitation focus. On OpenVLA-7b, \texttt{Sorting\_RF} improves TC by $+3.7$~pp over B1 and $+2.7$~pp over B2. On \pimodel, \texttt{Sorting\_DT} is highest at 84.2\%, exceeding both baselines. Notably, \texttt{FSCSART\_DT} shows a slight TC dip on EO1 (79.6\%) and \pimodel (80.2\%) relative to both baselines, indicating that hard-partition DT guidance can sometimes over-exploit failure regions at a small cost to diversity.
\paragraph{Trajectory Coverage under Failure (TCF):}
TCF measures how behaviorally diverse the discovered failures are. Detailed TCF statistics are summarized in Appendix~\ref{sec:appendix_tables}, Table~\ref{tab:tcf}, with distributions provided in Figure~\ref{fig:tcf_boxplots} (Appendix~\ref{sec:appendix_boxplots}). On GR00T-N1.6, \texttt{Sorting\_RF} achieves 83.0\% TCF, exceeding both B1 (78.2\%, $+4.8$~pp) and B2 (79.9\%, $+3.1$~pp). On \pimodel, \texttt{Sorting\_RF} reaches 77.4\% vs.\ 72.1\% for B1 ($+5.3$~pp) and 72.6\% for B2 ($+4.8$~pp). On OpenVLA-7b, \texttt{Sorting\_RF} and \texttt{Weighted\_RF} both achieve 70.1\% vs.\ 63.2\% for B1 ($+6.9$~pp) and 65.4\% for B2 ($+4.7$~pp). The only exception is EO1, where \texttt{FSCSART\_DT} falls to 67.2\% — below both B1 (70.8\%) and B2 (70.3\%) — while \texttt{Sorting\_RF} and \texttt{Sorting\_DT} recover to 71.3\% and 72.2\% respectively, surpassing both baselines. This confirms that the DT alone can over-focus failures into a narrow behavioral region, whereas RF-based and sorted variants maintain broader failure diversity.
\paragraph{Failed Object Coverage (FOC):}
FOC measures the ratio of objects found to fail for each method. Detailed FOC statistics are reported in Appendix~\ref{sec:appendix_tables}, Table~\ref{tab:foc}, with distributions shown in Figure~\ref{fig:foc_boxplots} (Appendix~\ref{sec:appendix_boxplots}). Due to the simplicity of uncovering failures in OpenVLA-7b, all methods saturate at 100\% FOC. On \pimodel, \texttt{Sorting\_RF} and \texttt{Weighted\_RF} both achieve 100\% FOC, matching B2 and improving over B1's 95.7\% ($+4.3$~pp). On GR00T-N1.6, \texttt{Weighted\_RF} (97.1\%) exceeds both B1 (95.7\%, $+1.4$~pp) and B2 (95.8\%, $+1.3$~pp) while simultaneously delivering a $+27.2$~pp FR gain. A notable pattern is that \texttt{Sorting\_RF} shows lower FOC on EO1 (92.9\%, below B1 and B2) and GR00T-N1.6 (84.8\%), indicating that sorted RF selection can over-concentrate failures on high-probability objects. \texttt{Weighted\_RF} avoids this by blending failure probability with a diversity normalization term, consistently matching or exceeding both baselines in FOC.


\section{RQ2: Impact of the ML Model (DT vs.\ RF) on Failure Detection}
\label{sec:appendix_rq2_details}
\label{sec:rq2}

RQ2 isolates the contribution of the machine-learning guidance component, and compares the effect of using a Decision Tree (DT) vs.\ a Random Forest (RF) on both the quantity and diversity of discovered failures.

\paragraph{DT contribution.}
Comparing \texttt{FSCSART\_DT} against B2:~FSCSART directly isolates the DT's effect, since both methods use the same ART diversity mechanism and differ only in whether a learned failure model is applied. The gains are large: $+18.6$~pp on EO1 (38.0\% $\to$ 56.6\%), $+28.3$~pp on GR00T-N1.6 (36.5\% $\to$ 64.8\%), and $+13.3$~pp on \pimodel (55.5\% $\to$ 68.8\%). The DT learns a partition of the scene feature space that separates passing from failing configurations, redirecting candidate selection towards failure-dense input regions that a pure diversity strategy never reaches.

\paragraph{RF vs.\ DT.}
Comparing RF-based methods (\texttt{Sorting\_RF}, \texttt{Weighted\_RF}) against their DT counterparts (\texttt{Sorting\_DT}, \texttt{FSCSART\_DT}) reveals two consistent patterns. First, on FR, the RF variants are broadly competitive: on EO1, \texttt{Sorting\_RF} (60.0\%) slightly edges \texttt{Sorting\_DT} (52.7\%); on \pimodel, \texttt{Weighted\_RF} (73.7\%) outperforms both DT variants. Second, and more importantly, RF variants achieve substantially higher TCF (Table~\ref{tab:tcf}) than the DT-only variant. On \pimodel, \texttt{Sorting\_RF} (77.4\%) exceeds \texttt{FSCSART\_DT} (68.6\%) by 8.8~pp while matching its FR; on GR00T-N1.6, \texttt{Sorting\_RF} (83.0\%) exceeds \texttt{FSCSART\_DT} (81.0\%) while also leading on FR. This reflects the fundamental difference between the two model types: a DT partitions the feature space with hard boundaries, which can over-concentrate test generation on a single dominant failure cluster (high FR, low TCF), while an RF averages across multiple trees to produce smoother failure-probability estimates that preserve broader exploration of the failure space (high FR, high TCF). The \texttt{Weighted\_RF} variant further enhances this by blending the RF failure probability with a normalised diversity score, yielding the best overall balance across FR, TCF, and FOC.


\section{Detailed Experimental Tables}
\label{sec:appendix_tables}

\begin{table}[!ht]
\centering
\caption{Median Failure Rate (\%). Baselines separated from proposed methods by a vertical rule. $\Delta$B1 and $\Delta$B2 columns show the improvement of the best proposed method over each baseline. Best proposed result per row in \textbf{bold} and \underline{underlined}.}
\label{tab:fr}

\resizebox{\linewidth}{!}{%
\begin{tabular}{lcc|cccccc}
\toprule
\textbf{VLA} & \textbf{B1: Random} & \textbf{B2: FSCSART}
& \textbf{FSCSART\_DT} & \textbf{Sorting\_RF}
& \textbf{Sorting\_DT} & \textbf{Weighted\_RF}
& \textbf{$\Delta$B1 (\%)} & \textbf{$\Delta$B2 (\%)} \\
\cmidrule{1-9}
EO1         & 36.7 & 38.0 & 56.6 & \underline{\textbf{60.0}} & 52.7 & 59.6 & +23.3 & +22.0 \\
GR00T-N1.6 & 35.6 & 36.5 & 64.8 & \underline{\textbf{65.3}} & 62.8 & 62.8 & +29.7 & +28.8 \\
OpenVLA-7b & 95.3 & 95.7 & 96.5 & 97.0 & 96.8 & \underline{\textbf{97.5}} & +2.2 & +1.8 \\
\pimodel    & 54.0 & 55.5 & 68.8 & 68.8 & 65.1 & \underline{\textbf{73.7}} & +19.7 & +18.2 \\
\bottomrule
\end{tabular}}
\end{table}

\begin{table}[!ht]
\centering
\caption{Median Trajectory Coverage, TC (\%). Baselines separated from proposed methods by a vertical rule. $\Delta$B1 and $\Delta$B2 columns show the improvement of the best proposed method over each baseline. Best proposed result per row in \textbf{bold} and \underline{underlined}.}
\label{tab:tc}

\resizebox{\linewidth}{!}{%
\begin{tabular}{lcc|cccccc}
\toprule
\textbf{VLA} & \textbf{B1: Random} & \textbf{B2: FSCSART}
& \textbf{FSCSART\_DT} & \textbf{Sorting\_RF}
& \textbf{Sorting\_DT} & \textbf{Weighted\_RF}
& \textbf{$\Delta$B1 (\%)} & \textbf{$\Delta$B2 (\%)} \\
\cmidrule{1-9}
EO1         & 82.8 & 81.5 & 79.6 & 81.3 & \underline{\textbf{84.0}} & 83.1 & +1.2 & +2.5 \\
GR00T-N1.6 & \underline{\textbf{$\sim$86}} & \underline{\textbf{$\sim$86}} & \underline{\textbf{$\sim$86}} & \underline{$\sim$\textbf{86}} & 78.1 & \underline{$\sim$\textbf{86}} & +0.0 & +0.0 \\
OpenVLA-7b & 66.8 & 67.8 & 66.5 & \underline{\textbf{70.5}} & 68.3 & 70.1 & +3.7 & +2.7 \\
\pimodel    & 82.8 & 83.2 & 80.2 & 82.8 & \underline{\textbf{84.2}} & 83.1 & +1.4 & +1.0 \\
\bottomrule
\end{tabular}}
\end{table}

\begin{table}[!ht]
\centering
\caption{Median Trajectory Coverage under Failure, TCF (\%). Baselines separated from proposed methods by a vertical rule. $\Delta$B1 and $\Delta$B2 columns show the improvement of the best proposed method over each baseline. Best proposed result per row in \textbf{bold} and \underline{underlined}.}
\label{tab:tcf}

\resizebox{\linewidth}{!}{%
\begin{tabular}{lcc|cccccc}
\toprule
\textbf{VLA} & \textbf{B1: Random} & \textbf{B2: FSCSART}
& \textbf{FSCSART\_DT} & \textbf{Sorting\_RF}
& \textbf{Sorting\_DT} & \textbf{Weighted\_RF}
& \textbf{$\Delta$B1 (\%)} & \textbf{$\Delta$B2 (\%)} \\
\cmidrule{1-9}
EO1         & 70.8 & 70.3 & 67.2 & 71.3 & \underline{\textbf{72.2}} & 71.1 & +1.4 & +1.9 \\
GR00T-N1.6 & 78.2 & 79.9 & 81.0 & \underline{\textbf{83.0}} & 81.6 & 82.5 & +4.8 & +3.1 \\
OpenVLA-7b & 63.2 & 65.4 & 66.5 & \underline{\textbf{70.1}} & 66.8 & \underline{\textbf{70.1}} & +6.9 & +4.7 \\
\pimodel    & 72.1 & 72.6 & 68.6 & \underline{\textbf{77.4}} & 76.2 & 72.7 & +5.3 & +4.8 \\
\bottomrule
\end{tabular}}
\end{table}

\begin{table}[!ht]
\centering
\caption{Median Failed Object Coverage, FOC (\%). Baselines separated from proposed methods by a vertical rule. $\Delta$B1 and $\Delta$B2 columns show the improvement of the best proposed method over each baseline. Best proposed result per row in \textbf{bold} and \underline{underlined}.}
\label{tab:foc}

\resizebox{\linewidth}{!}{%
\begin{tabular}{lcc|cccccc}
\toprule
\textbf{VLA} & \textbf{B1: Random} & \textbf{B2: FSCSART}
& \textbf{FSCSART\_DT} & \textbf{Sorting\_RF}
& \textbf{Sorting\_DT} & \textbf{Weighted\_RF}
& \textbf{$\Delta$B1 (\%)} & \textbf{$\Delta$B2 (\%)} \\
\cmidrule{1-9}
EO1         & 96.1 & 98.6 & 95.7 & 92.9 & 97.1 & \underline{\textbf{98.6}} & +2.5 & +0.0 \\
GR00T-N1.6 & 95.7 & 95.8 & 88.6 & 84.8 & 92.9 & \underline{\textbf{97.1}} & +1.4 & +1.3 \\
OpenVLA-7b & \underline{\textbf{100.0}} & \underline{\textbf{100.0}} & \underline{\textbf{100.0}}
            & \underline{\textbf{100.0}} & \underline{\textbf{100.0}} & 98.6 & +0.0 & +0.0 \\
\pimodel    & 95.7 & \underline{\textbf{100.0}} & 97.1 & \underline{\textbf{100.0}} & 98.6 & \underline{\textbf{100.0}} & +4.3 & +0.0 \\
\bottomrule
\end{tabular}}
\end{table}

\FloatBarrier

\section{Box Plot Distributions for All Metrics}
\label{sec:appendix_boxplots}

This appendix provides the full interquartile-range box plots for all four evaluation metrics — Failure Rate (FR), Trajectory Coverage (TC), Trajectory Coverage under Failure (TCF), and Failed Object Coverage (FOC) — across the four VLA models. Annotated values in each plot are medians over 10 independent runs.

%% file: references.bib
@misc{black2024pi0visionlanguageactionflowmodel,
      title={$\pi_0$: A Vision-Language-Action Flow Model for General Robot Control}, 
      author={Kevin Black and Noah Brown and Danny Driess and Adnan Esmail and Michael Equi and Chelsea Finn and Niccolo Fusai and Lachy Groom and Karol Hausman and Brian Ichter and Szymon Jakubczak and Tim Jones and Liyiming Ke and Sergey Levine and Adrian Li-Bell and Mohith Mothukuri and Suraj Nair and Karl Pertsch and Lucy Xiaoyang Shi and James Tanner and Quan Vuong and Anna Walling and Haohuan Wang and Ury Zhilinsky},
      year={2024},
      eprint={2410.24164},
      archivePrefix={arXiv},
      primaryClass={cs.LG},
      note = {\url{https://arxiv.org/abs/2410.24164}}
}

@article{valle2025evaluating,
  title={Evaluating uncertainty and quality of visual language action-enabled robots},
  author={Valle, Pablo and Lu, Chengjie and Ali, Shaukat and Arrieta, Aitor},
  journal={arXiv preprint arXiv:2507.17049},
  year={2025}
}

@misc{peng2025nebula,
      title={NEBULA: Do We Evaluate Vision-Language-Action Agents Correctly?}, 
      author={Jierui Peng and Yanyan Zhang and Yicheng Duan and Tuo Liang and Vipin Chaudhary and Yu Yin},
      year={2025},
      eprint={2510.16263},
      archivePrefix={arXiv},
      primaryClass={cs.RO},
      note = {\url{https://arxiv.org/abs/2510.16263}}
}

@misc{qu2025eo1,
      title={EO-1: Interleaved Vision-Text-Action Pretraining for General Robot Control}, 
      author={Delin Qu and Haoming Song and Qizhi Chen and Zhaoqing Chen and Xianqiang Gao and Xinyi Ye and Qi Lv and Modi Shi and Guanghui Ren and Cheng Ruan and Maoqing Yao and Haoran Yang and Jiacheng Bao and Bin Zhao and Dong Wang},
      year={2025},
      eprint={2508.21112},
      archivePrefix={arXiv},
      primaryClass={cs.RO},
      note = {\url{https://arxiv.org/abs/2508.21112}} 
}

@article{li24simpler,
         title={Evaluating Real-World Robot Manipulation Policies in Simulation},
         author={Xuanlin Li and Kyle Hsu and Jiayuan Gu and Karl Pertsch and Oier Mees and Homer Rich Walke and Chuyuan Fu and Ishikaa Lunawat and Isabel Sieh and Sean Kirmani and Sergey Levine and Jiajun Wu and Chelsea Finn and Hao Su and Quan Vuong and Ted Xiao},
         journal = {arXiv preprint arXiv:2405.05941},
         year={2024}
}

@inproceedings{zhang2025vlabench,
  title={Vlabench: A large-scale benchmark for language-conditioned robotics manipulation with long-horizon reasoning tasks},
  author={Zhang, Shiduo and Xu, Zhe and Liu, Peiju and Yu, Xiaopeng and Li, Yuan and Gao, Qinghui and Fei, Zhaoye and Yin, Zhangyue and Wu, Zuxuan and Jiang, Yu-Gang and others},
  booktitle={Proceedings of the IEEE/CVF International Conference on Computer Vision},
  pages={11142--11152},
  year={2025}
}

@misc{nvidia2025gr00tn1openfoundation,
      title={GR00T N1: An Open Foundation Model for Generalist Humanoid Robots}, 
      author={NVIDIA and : and Johan Bjorck and Fernando Castañeda and Nikita Cherniadev and Xingye Da and Runyu Ding and Linxi "Jim" Fan and Yu Fang and Dieter Fox and Fengyuan Hu and Spencer Huang and Joel Jang and Zhenyu Jiang and Jan Kautz and Kaushil Kundalia and Lawrence Lao and Zhiqi Li and Zongyu Lin and Kevin Lin and Guilin Liu and Edith Llontop and Loic Magne and Ajay Mandlekar and Avnish Narayan and Soroush Nasiriany and Scott Reed and You Liang Tan and Guanzhi Wang and Zu Wang and Jing Wang and Qi Wang and Jiannan Xiang and Yuqi Xie and Yinzhen Xu and Zhenjia Xu and Seonghyeon Ye and Zhiding Yu and Ao Zhang and Hao Zhang and Yizhou Zhao and Ruijie Zheng and Yuke Zhu},
      year={2025},
      eprint={2503.14734},
      archivePrefix={arXiv},
      primaryClass={cs.RO},
      note = {\url{https://arxiv.org/abs/2503.14734}}
}

@inproceedings{openx2023,
  title={Open x-embodiment: Robotic learning datasets and rt-x models: Open x-embodiment collaboration 0},
  author={O’Neill, Abby and Rehman, Abdul and Maddukuri, Abhiram and Gupta, Abhishek and Padalkar, Abhishek and Lee, Abraham and Pooley, Acorn and Gupta, Agrim and Mandlekar, Ajay and Jain, Ajinkya and others},
  booktitle={2024 IEEE International Conference on Robotics and Automation (ICRA)},
  pages={6892--6903},
  year={2024},
  organization={IEEE}
}

@article{calvin2022,
  title={Calvin: A benchmark for language-conditioned policy learning for long-horizon robot manipulation tasks},
  author={Mees, Oier and Hermann, Lukas and Rosete-Beas, Erick and Burgard, Wolfram},
  journal={IEEE Robotics and Automation Letters},
  volume={7},
  number={3},
  pages={7327--7334},
  year={2022},
  publisher={IEEE}
}

@article{libero2023,
  title={Libero: Benchmarking knowledge transfer for lifelong robot learning},
  author={Liu, Bo and Zhu, Yifeng and Gao, Chongkai and Feng, Yihao and Liu, Qiang and Zhu, Yuke and Stone, Peter},
  journal={Advances in Neural Information Processing Systems},
  volume={36},
  pages={44776--44791},
  year={2023}
}

@article{maniskill22023,
  title={Maniskill2: A unified benchmark for generalizable manipulation skills},
  author={Gu, Jiayuan and Xiang, Fanbo and Li, Xuanlin and Ling, Zhan and Liu, Xiqiang and Mu, Tongzhou and Tang, Yihe and Tao, Stone and Wei, Xinyue and Yao, Yunchao and others},
  journal={arXiv preprint arXiv:2302.04659},
  year={2023}
}

@article{vlatest2024,
  title={Vlatest: Testing and evaluating vision-language-action models for robotic manipulation},
  author={Wang, Zhijie and Zhou, Zhehua and Song, Jiayang and Huang, Yuheng and Shu, Zhan and Ma, Lei},
  journal={Proceedings of the ACM on Software Engineering},
  volume={2},
  number={FSE},
  pages={1615--1638},
  year={2025},
  publisher={ACM New York, NY, USA}
}

@inproceedings{valle2026metamorphic,
            title        = {Metamorphic Testing of Vision-Language Action-Enabled Robots},
            author       = {Pablo Valle and Sergio Segura and Shaukat Ali and Aitor Arrieta},
            booktitle    = {Proceedings of the 19th IEEE International Conference on Software Testing, Verification and Validation (ICST 2026)},
            year         = {2026},
            month        = may,
            address      = {Daejeon, Republic of Korea},
            organization = {IEEE},
            note = {\url{https://arxiv.org/abs/2602.22579}}
          }

@inproceedings{chen2004adaptive,
  title={Adaptive random testing},
  author={Chen, Tsong Yueh and Leung, Hing and Mak, Ieng Kei},
  booktitle={Annual Asian Computing Science Conference},
  pages={320--329},
  year={2004},
  organization={Springer}
}

@misc{FATE-VLA_code,
author = {Kanwal, Arusa and Valle, Pablo and Ali, Shaukat and Arrieta, Aitor},
  title = {Replication Package for paper FATE-VLA: Failure-Aware Test Generation for Vision-Language-Action Models},
  note = {\url{https://github.com/pablovalle/FATE-VLA}},
month = may,
year = 2026,
}

@article{brohan2022rt,
  title={Rt-1: Robotics transformer for real-world control at scale},
  author={Brohan, Anthony and Brown, Noah and Carbajal, Justice and Chebotar, Yevgen and Dabis, Joseph and Finn, Chelsea and Gopalakrishnan, Keerthana and Hausman, Karol and Herzog, Alex and Hsu, Jasmine and others},
  journal={arXiv preprint arXiv:2212.06817},
  year={2022}
}

@article{qu2025spatialvla,
  title={SpatialVLA: Exploring Spatial Representations for Visual-Language-Action Model},
  author={Qu, Delin and Song, Haoming and Chen, Qizhi and Yao, Yuanqi and Ye, Xinyi and Ding, Yan and Wang, Zhigang and Gu, JiaYuan and Zhao, Bin and Wang, Dong and others},
  journal={arXiv preprint arXiv:2501.15830},
  year={2025}
}

@article{wang2025vlatest,
  title={Vlatest: Testing and evaluating vision-language-action models for robotic manipulation},
  author={Wang, Zhijie and Zhou, Zhehua and Song, Jiayang and Huang, Yuheng and Shu, Zhan and Ma, Lei},
  journal={Proceedings of the ACM on Software Engineering},
  volume={2},
  number={FSE},
  pages={1615--1638},
  year={2025},
  publisher={ACM New York, NY, USA}
}

@article{kim2024openvla,
  title={Openvla: An open-source vision-language-action model},
  author={Kim, Moo Jin and Pertsch, Karl and Karamcheti, Siddharth and Xiao, Ted and Balakrishna, Ashwin and Nair, Suraj and Rafailov, Rafael and Foster, Ethan and Lam, Grace and Sanketi, Pannag and others},
  journal={arXiv preprint arXiv:2406.09246},
  year={2024}
}

@article{liu2025eva,
  title={Eva-VLA: Evaluating Vision-Language-Action Models' Robustness Under Real-World Physical Variations},
  author={Liu, Hanqing and Ruan, Shouwei and Long, Jiahuan and Wu, Junqi and Hou, Jiacheng and Tang, Huili and Jiang, Tingsong and Zhou, Weien and Yao, Wen},
  journal={arXiv preprint arXiv:2509.18953},
  year={2025}
}

@article{tong2026uncovering,
  title={Uncovering Linguistic Fragility in Vision-Language-Action Models via Diversity-Aware Red Teaming},
  author={Tong, Baoshun and He, Haoran and Pan, Ling and Liu, Yang and Lin, Liang},
  journal={arXiv preprint arXiv:2604.05595},
  year={2026}
}

@article{fei2025libero,
  title={Libero-plus: In-depth robustness analysis of vision-language-action models},
  author={Fei, Senyu and Wang, Siyin and Shi, Junhao and Dai, Zihao and Cai, Jikun and Qian, Pengfang and Ji, Li and He, Xinzhe and Zhang, Shiduo and Fei, Zhaoye and others},
  journal={arXiv preprint arXiv:2510.13626},
  year={2025}
}

@article{wang2026libero,
  title={LIBERO-X: Robustness Litmus for Vision-Language-Action Models},
  author={Wang, Guodong and Zhang, Chenkai and Liu, Qingjie and Zhang, Jinjin and Cai, Jiancheng and Liu, Junjie and Liu, Xinmin},
  journal={arXiv preprint arXiv:2602.06556},
  year={2026}
}

@article{gu2026safe,
  title={Safe: Multitask failure detection for vision-language-action models},
  author={Gu, Qiao and Ju, Yuanliang and Sun, Shengxiang and Gilitschenski, Igor and Nishimura, Haruki and Itkina, Masha and Shkurti, Florian},
  journal={Advances in Neural Information Processing Systems},
  volume={38},
  pages={40041--40076},
  year={2026}
}

@article{lee2020adaptive,
  title={Adaptive stress testing: Finding likely failure events with reinforcement learning},
  author={Lee, Ritchie and Mengshoel, Ole J and Saksena, Anshu and Gardner, Ryan W and Genin, Daniel and Silbermann, Joshua and Owen, Michael and Kochenderfer, Mykel J},
  journal={Journal of Artificial Intelligence Research},
  volume={69},
  pages={1165--1201},
  year={2020}
}

@inproceedings{haq2022efficient,
  title={Efficient online testing for DNN-enabled systems using surrogate-assisted and many-objective optimization},
  author={Haq, Fitash Ul and Shin, Donghwan and Briand, Lionel},
  booktitle={Proceedings of the 44th international conference on software engineering},
  pages={811--822},
  year={2022}
}

@article{pumacay2024colosseum,
  title={The colosseum: A benchmark for evaluating generalization for robotic manipulation},
  author={Pumacay, Wilbert and Singh, Ishika and Duan, Jiafei and Krishna, Ranjay and Thomason, Jesse and Fox, Dieter},
  journal={arXiv preprint arXiv:2402.08191},
  year={2024}
}

@inproceedings{sheikhi2022coverage,
  title={Coverage-guided fuzz testing for cyber-physical systems},
  author={Sheikhi, Sanaz and Kim, Edward and Duggirala, Parasara Sridhar and Bak, Stanley},
  booktitle={2022 ACM/IEEE 13th International Conference on Cyber-Physical Systems (ICCPS)},
  pages={24--33},
  year={2022},
  organization={IEEE}
}

@article{atreya2025roboarena,
  title={Roboarena: Distributed real-world evaluation of generalist robot policies},
  author={Atreya, Pranav and Pertsch, Karl and Lee, Tony and Kim, Moo Jin and Jain, Arhan and Kuramshin, Artur and Eppner, Clemens and Neary, Cyrus and Hu, Edward and Ramos, Fabio and others},
  journal={arXiv preprint arXiv:2506.18123},
  year={2025}
}

@article{farid2022failure,
  title={Failure prediction with statistical guarantees for vision-based robot control},
  author={Farid, Alec and Snyder, David and Ren, Allen Z and Majumdar, Anirudha},
  journal={arXiv preprint arXiv:2202.05894},
  year={2022}
}
